\documentclass{article}
\usepackage[T1]{fontenc}
\usepackage{enumitem}
\usepackage{iclr2027_conference,times}
\newif\ificlrversion
\iclrversiontrue
\iclrfinalcopy
\usepackage{amsmath}
\usepackage{amssymb}
\usepackage{amsthm}
\usepackage{mathtools}
\usepackage{graphicx}
\usepackage{booktabs}
\usepackage{multirow}
\usepackage{tabularx}
\usepackage{wrapfig}
\usepackage{algorithm}
\usepackage{algorithmic}
\usepackage{xcolor}
\usepackage{microtype}
\usepackage{placeins}
\usepackage{url}
\usepackage{hyperref}
\usepackage{tikz}
\usetikzlibrary{arrows.meta,positioning,fit,calc,backgrounds}

\newif\ificlrrepro

\definecolor{enpdablue}{HTML}{1769AA}
\definecolor{enpdateal}{HTML}{087E8B}
\definecolor{enpdared}{HTML}{C73E3A}
\definecolor{enpdagray}{HTML}{F1F3F5}
\hypersetup{
  colorlinks=true,
  citecolor=enpdablue,
  linkcolor=enpdablue,
  urlcolor=enpdablue
}

\newcommand{\vect}[1]{\operatorname{vec}(#1)}

\newcommand{\one}{\mathbf{1}}
\newcommand{\soft}{\mathbf{S}}
\newcommand{\hard}{\mathbf{P}}
\newcommand{\aff}{\mathbf{A}}
\newcommand{\price}{\boldsymbol{\lambda}}
\newcommand{\bid}{\mathbf{Z}}

\DeclareMathOperator{\diag}{diag}

\newtheorem{theorem}{Theorem}
\newtheorem{proposition}[theorem]{Proposition}

\theoremstyle{definition}

\theoremstyle{remark}

\newcommand{\confint}[2]{{\footnotesize[#1,\,#2]}}

\title{{\fontsize{15}{18}\selectfont Equivariant Neural Primal--Dual Assignment for\\[-0.5mm]
Maximum Common Edge Subgraphs}}

\author{%
\begin{tabular}{@{}*{3}{p{\dimexpr(\textwidth-2\tabcolsep)/3\relax}@{}}}
\textbf{Jiaqing Xie}\textsuperscript{1,2,3} &
\textbf{Yanchao Li}\textsuperscript{4} &
\textbf{Zhuo Yang}\textsuperscript{5,3} \\[0.7ex]
\textbf{Yuxin Wang}\textsuperscript{3} &
\textbf{Tianfan Fu}\textsuperscript{4,}\thanks{%
Corresponding authors: Tianfan Fu and Yuqiang Li.\\
Emails: \href{mailto:26113050148@m.fudan.edu.cn}{\nolinkurl{26113050148@m.fudan.edu.cn}} (Jiaqing Xie);\\
\href{mailto:futianfan@gmail.com}{\nolinkurl{futianfan@gmail.com}} (Tianfan Fu);
\href{mailto:liyuqiang@pjlab.org.cn}{\nolinkurl{liyuqiang@pjlab.org.cn}} (Yuqiang Li).\\
Code: \url{https://github.com/jiaqingxie/ENPDA-MCES}.} &
\textbf{Yuqiang Li}\textsuperscript{2,}\footnotemark[1]
\end{tabular}\\[1.2ex]
{\normalfont\small\textsuperscript{1}Fudan University\quad
\textsuperscript{2}Shanghai AI Lab\quad
\textsuperscript{3}Shanghai Innovation Institute}\\
{\normalfont\small\textsuperscript{4}Nanjing University\quad
\textsuperscript{5}Southeast University}
}

\begin{document}
\maketitle
\begin{abstract}
Maximum common edge subgraph (MCES) matching finds a partial vertex correspondence between two labeled graphs that preserves as many labeled edges as possible. Molecular similarity search requires matching many graph pairs, making the cost of repeated queries important. The strongest baseline attains accurate MCES solutions but trains a separate network for each pair. We introduce \textit{Equivariant Neural Primal-Dual Assignment \mbox{(ENPDA)}}, which learns a shared matching policy and applies it to new pairs without further training, answering queries roughly three orders of magnitude faster and recovering its training cost after a few dozen queries. The policy recomputes exact objective marginals for candidate matches and learns corrections and step sizes that update their scores. Target prices respond to competition when several source vertices favor the same target. Four update rounds and a Hungarian projection produce a partial one-to-one matching. We prove per-pair guarantees that hold for any network parameters. In exact arithmetic, reordering either graph permutes the assignment and price states, the projected matching is one-to-one, and repaired prices give a valid MCES upper bound. Subtracting the preserved-edge count bounds the optimality gap; combined with structural caps, these certificates prove global optimality for 60 of 291 native test pairs. On three molecular benchmarks with disjoint train/validation/test splits, ENPDA improves over an analytic counterpart with the same update and projection budget by 7.4-8.6 accuracy points; after one second of refinement search, 2.5-3 points of the gain remain. Transferred without fine-tuning to edge-deletion tasks from social and protein graphs, the policy gains 9.1-17.6 points over the analytic counterpart. When output matchings must keep aromatic rings intact, ENPDA recovers more reference bonds than the baselines on all three datasets.
\end{abstract}

\section{Introduction}
\label{sec:introduction}

Maximum common edge subgraph (MCES) matching finds, between two labeled graphs, a partial
vertex correspondence that preserves as many identically labeled edges as possible.
Applications include similarity search and lead optimization in drug
discovery~\citep{raymond2002heuristics,ehrlich2011maximum}, pattern
recognition~\citep{conte2004thirty,foggia2014graph}, and bioinformatics network
alignment~\citep{singh2008global}. Molecular correspondences can be inspected atom by atom
and bond by bond. Finding an optimum is
NP-hard~\citep{garey1979computers,bunke1998graph,ehrlich2011maximum}.

Classical MCES solvers use maximum clique search or integer programming
formulations~\citep{raymond2002rascal,degastines2024formulations}, which can require substantial
computation on difficult graph pairs.  Graph neural networks (GNNs) learn structural information
across pairs for graph-similarity prediction and node matching~\citep{li2019gmn,wang2021ngm}.
For MCES, Neural Graduated Assignment (NGA)~\citep{ying2026neural} obtains strong approximate
solutions by optimizing a network over an association common graph (ACG) separately for every
queried pair.  Repeated queries therefore incur repeated training.  We ask whether the MCES
update policy can instead be learned across pairs and frozen at inference, letting
instance-specific assignment states adapt to competing matches under one-to-one mapping
constraints.

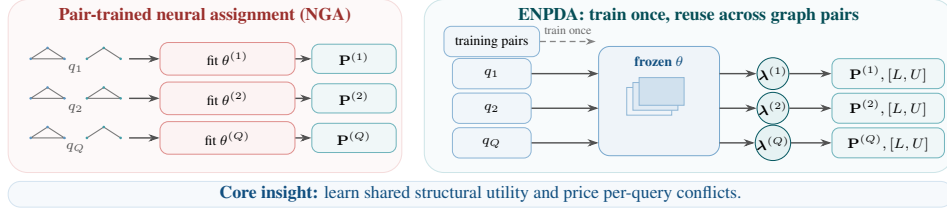
\begin{figure*}[t]
  \centering
  \resizebox{.90\textwidth}{!}{\begin{tikzpicture}[
  x=1mm,y=.88mm,
  flow/.style={-{Stealth[length=2mm,width=1.35mm]},draw=black!68,line width=.72pt},
  softflow/.style={-{Stealth[length=1.8mm,width=1.2mm]},draw=black!48,
    densely dashed,line width=.65pt},
  fitbox/.style={draw=enpdared!58,fill=enpdared!6,rounded corners=1.5mm,
    minimum width=24mm,minimum height=5.8mm,align=center,font=\scriptsize},
  outbox/.style={draw=enpdateal!58,fill=enpdateal!6,rounded corners=1.3mm,
    minimum width=15mm,minimum height=5.2mm,align=center,font=\scriptsize},
  trainbox/.style={draw=enpdablue!50,fill=enpdablue!6,rounded corners=1.3mm,
    minimum width=17mm,minimum height=5.4mm,align=center,font=\scriptsize},
  price/.style={circle,draw=enpdateal!70!black,fill=enpdateal!10,
    minimum size=5.4mm,inner sep=0pt,font=\scriptsize},
]
  % Two visual regimes and one shared thesis strip.
  \filldraw[draw=enpdared!18,fill=enpdared!3,rounded corners=3.5mm]
    (0,7) rectangle (70,42);
  \filldraw[draw=enpdateal!20,fill=enpdateal!3,rounded corners=3.5mm]
    (74,7) rectangle (168,42);
  \filldraw[draw=enpdablue!18,fill=enpdablue!4,rounded corners=2.2mm]
    (0,0) rectangle (168,5.2);

  \node[font=\small\bfseries,text=enpdared!78!black] at (35,38.8)
    {Pair-trained neural assignment (NGA)};

  % Compact graph-pair glyphs.  Their topology changes slightly by row, while
  % node ordering remains intentionally irrelevant.
  \foreach \yy/\lab/\extra in {30/$q_1$/0,22/$q_2$/1,14/$q_Q$/0}{
    \begin{scope}[shift={(4,\yy)}]
      \draw[black!52,line width=.55pt] (0,0)--(3,2.2)--(6,0)--(0,0);
      \fill[enpdablue!70] (0,0) circle (.65pt) (3,2.2) circle (.65pt) (6,0) circle (.65pt);
      \draw[black!52,line width=.55pt] (10,0)--(13,2.2)--(16,0);
      \ifnum\extra=1 \draw[black!52,line width=.55pt] (10,0)--(16,0);\fi
      \fill[enpdateal!76] (10,0) circle (.65pt) (13,2.2) circle (.65pt) (16,0) circle (.65pt);
      \node[font=\scriptsize,text=black!68] at (8,-2.0) {\lab};
    \end{scope}
  }

  \node[fitbox] (fitone) at (39,30) {fit $\theta^{(1)}$};
  \node[fitbox] (fittwo) at (39,22) {fit $\theta^{(2)}$};
  \node[fitbox] (fitq)   at (39,14) {fit $\theta^{(Q)}$};
  \node[outbox] (pone) at (61.5,30) {$\hard^{(1)}$};
  \node[outbox] (ptwo) at (61.5,22) {$\hard^{(2)}$};
  \node[outbox] (pq)   at (61.5,14) {$\hard^{(Q)}$};
  \foreach \yy/\fit/\out in {30/fitone/pone,22/fittwo/ptwo,14/fitq/pq}{
    \draw[flow] (21.5,\yy)--(\fit.west);
    \draw[flow] (\fit.east)--(\out.west);
  }

  \node[font=\small\bfseries,text=enpdateal!62!black] at (121,38.8)
    {ENPDA: train once, reuse across graph pairs};

  % One-time training writes a single shared policy.
  \node[trainbox] (trainset) at (86,33.5) {training pairs};
  \draw[softflow] (trainset.east)--node[above,font=\tiny,text=black!58]
    {train once}(104.5,33.5);

  % Frozen shared policy with explicit representation planes.
  \filldraw[draw=enpdablue!58,fill=enpdablue!7,rounded corners=2mm]
    (105,10.4) rectangle (126.5,32.5);
  \node[font=\scriptsize\bfseries,text=enpdablue!75!black] at (115.75,29.6)
    {frozen $\theta$};
  \foreach \k/\tone in {0/7,1/13,2/20}{
    \draw[draw=enpdablue!48,fill=enpdablue!\tone]
      (110+1.1*\k,18.6+1.0*\k) rectangle +(8.0,5.6);
  }

  % Every query uses the same theta but carries its own target prices.
  \foreach \yy/\lab/\plab in {27/$q_1$/$\price^{(1)}$,20/$q_2$/$\price^{(2)}$,13/$q_Q$/$\price^{(Q)}$}{
    \node[trainbox,minimum width=14mm] (query-\yy) at (86,\yy) {\lab};
    \draw[flow] (query-\yy.east)--(105,\yy);
    \node[price] (price-\yy) at (136,\yy) {\plab};
    \draw[flow] (126.5,\yy)--(price-\yy.west);
    \node[outbox,minimum width=20mm] (result-\yy) at (156.5,\yy)
      {$\hard^{(\ifnum\yy=27 1\else\ifnum\yy=20 2\else Q\fi\fi)},[L,U]$};
    \draw[flow] (price-\yy.east)--(result-\yy.west);
  }

  \node[font=\small,text=enpdablue!75!black] at (84,2.6)
    {\textbf{Core insight:} learn shared structural utility and price per-query conflicts.};
\end{tikzpicture}}
  \caption{NGA fits each query separately~\citep{ying2026neural}.  ENPDA reuses a frozen
  equivariant policy, adapting bids and prices to return a matching and valid bounds.
  Symbols: $\theta$ denotes network parameters, $\price$ target prices, $\hard$ the
  returned matching, and $[L,U]$ the preserved-edge lower bound and valid upper bound
  (Section~\ref{sec:method}).}
  \label{fig:amortization-insight}
  \vspace{-1.5mm}
\end{figure*}

We introduce \textit{Equivariant Neural Primal--Dual Assignment (ENPDA)}, a neural auction
inspired by classical assignment algorithms~\citep{bertsekas2023auction}.  Source vertices act
as buyers, targets as objects, and bids express matching preferences.  A shared
equivariant network uses ACG structure to initialize bids and guide their updates.  When several
sources demand one target, its price rises and discourages that match.  The policy is frozen,
while bids and prices adapt to each pair
(Figure~\ref{fig:amortization-insight}).  A final Hungarian projection returns a partial
one-to-one correspondence.  Its preserved-edge count is a feasible lower bound, and repairing
prices against a linear assignment relaxation gives a valid upper bound.  Their difference gives an upper bound
on the optimality gap for any network parameters.

We evaluate the contribution of learning before and after search. \emph{ENPDA-Core}
applies four update rounds and one projection, and its output serves as the proposal
refined by \emph{ENPDA-Solver} through a fixed search portfolio. An analytic
counterpart replaces the learned components with fixed rules under the same budget,
isolating the effect of learning on final quality and runtime.

The main contributions of this paper are:
\begin{itemize}[leftmargin=*,parsep=0pt]
  \setlength{\itemsep}{1pt}
  \setlength{\topsep}{2pt}
  \item \textbf{A reusable equivariant matching policy for MCES.}  We introduce \textit{ENPDA}, which
  combines exact objective marginals with learned corrections and capacity-price updates to
  process new graph pairs without per-pair training, recovering its training cost after about
  40 queries against three NGA fits per query.
  \item \textbf{Per-pair guarantees and optimality bounds.}  For any network parameters,
  we prove in exact arithmetic that the continuous updates are equivariant under independent vertex relabelings,
  projection enforces one-to-one matching, and repaired prices yield valid upper bounds
  on the optimality gap of each returned matching; combined with structural and exact
  incident-assignment caps, these certificates prove global optimality for 60 of 291 native test pairs.
  \item \textbf{Controlled evidence for learning and transfer.}  With the same four rounds and
  one Hungarian projection, \textit{ENPDA-Core} improves over its analytic counterpart by 7.4--8.6
  accuracy points on three molecular benchmarks with graph-disjoint training, validation, and test
  splits.  Without fine-tuning, the same frozen checkpoints gain
  9.1--17.6 points over the analytic counterpart on 500 controlled
  edge-deletion pairs constructed from social and protein graphs.  \textit{ENPDA-Core} averages 0.19
  seconds per query versus 164--170 seconds for three NGA fits, and its proposals remain
  stronger after one second of additional search under the same budget.  When output matchings
  must keep aromatic rings intact, ENPDA-Solver recovers more reference bonds than RASCAL and
  NGA on all three datasets.
\end{itemize}

\section{Related work}
\label{sec:related}

\noindent\textbf{Maximum common subgraphs.}
Classical common-subgraph solvers use maximum-clique search, constraint programming, or
mixed-integer optimization~\citep{raymond2002heuristics,raymond2002rascal,ndiaye2011cp,
bahiense2012maximum,mccreesh2017mcsplit,degastines2024formulations}.  RASCAL searches molecular
modular products for maximum cliques~\citep{raymond2002rascal}.

\noindent\textbf{Neural graph matching and similarity.}
SimGNN~\citep{bai2019simgnn}, Graph Matching Networks (GMN)~\citep{li2019gmn},
NeuroMatch~\citep{lou2020neuromatch}, ISONET~\citep{roy2022isonet},
XMCS~\citep{roy2022maximum}, and INFMCS~\citep{lan2024interpretable} learn graph-level
similarities or latent correspondences.  Neural Graph Matching Network
(NGM)~\citep{wang2021ngm}, GANN-GM~\citep{wang2023unsupervised},
QC-DGM~\citep{gao2021quadratic}, and EQAN~\citep{tan2024ensemble} learn affinities or
association-graph updates.  Differentiable
assignment commonly uses Sinkhorn or Gumbel--Sinkhorn~\citep{sinkhorn1967concerning,
benamou2015iterative,mena2018gumbel}.  NGA~\citep{ying2026neural} is the closest MCES-specific
neural comparator, as it learns a graduated assignment procedure separately for each test pair.  ENPDA
instead learns one fixed-depth matching policy across pairs and makes conflicts explicit through dual
prices.

\noindent\textbf{Learning combinatorial algorithms.}
Learned optimizers amortize repeated numerical structure~\citep{andrychowicz2016learning}.  Neural
algorithmic reasoning trains graph networks to execute reusable
algorithms~\citep{velickovic2021neural}.  Primal--Dual Neural
Algorithmic Reasoning (PD-NAR)~\citep{he2025primaldual} trains GNNs to execute primal--dual
approximation algorithms for covering problems, using optimal small-instance solutions as auxiliary
supervision.  ENPDA instead tackles MCES, an indefinite quadratic assignment for which no known
primal--dual approximation trajectory is available to imitate, and learns its assignment updates
from exact ACG marginals and objective-generated training mappings.  Equivariant combinatorial
layers can retain an algorithmic approximation structure while learning across
instances~\citep{parmentier2025structured}.

\noindent\textbf{Auctions, assignment duality, and bounds.}
Auction algorithms interpret assignment dual variables as object prices and the Hungarian method
recovers an integral optimum for linear assignment~\citep{bertsekas2023auction}.  Our nonlinear
ACG utility prevents those classical guarantees from applying directly to MCES.  We use the auction
interpretation for reusable dynamics, then derive a separate linear incident-signature relaxation
whose repaired prices are rigorously dual feasible.  For tighter bounds, a sparse McCormick lifting
provides a conventional linear programming (LP) or mixed-integer linear programming (MILP)
upper bound~\citep{mccormick1976computability,
degastines2024formulations}.

\section{Equivariant Neural Primal--Dual Assignment}
\label{sec:method}

Each ENPDA round combines the graph-objective gradient, a learned structural correction,
and target prices that respond to competing assignments. A final projection enforces
one-to-one matching. Network weights are shared across pairs, while scores and prices reset for each query.

\subsection{Matching objective and association common graph}
Let $G_a=(V_a,E_a,\ell_a,\xi_a)$ and $G_b=(V_b,E_b,\ell_b,\xi_b)$ be simple undirected
node- and edge-labeled graphs, where $V$ collects vertices, $E$ edges, $\ell$ node labels,
and $\xi$ edge labels.  We orient the pair so $1\le n=|V_a|\le m=|V_b|$, \textit{i.e.}, the source graph
$G_a$ is the smaller one.  A partial injective map
$p:V_a\rightarrow V_b\cup\{\bot\}$, where $\bot$ marks an unmatched source vertex, preserves edge
$(i,k)\in E_a$ when $p(i),p(k)\neq\bot$, node labels agree, and
$(p(i),p(k))\in E_b$ has the same edge label.  MCES maximizes the size $|C(p)|$ of the
preserved-edge set $C(p)$.  Its binary assignment matrix $\hard$ has $P_{ij}=1$ exactly when
$p(i)=j$, with at most one nonzero per row and column.
The common subgraph need not be connected or induced, since extra edges in either input are allowed.

Each ACG vertex $(i,j)$ represents a candidate correspondence $i\mapsto j$.
We retain the full $n\times m$ grid and let $\mathcal C$ contain the node-label-compatible pairs.
An ACG edge joins compatible candidates $(i,j)$ and $(k,l)$ when $(i,k)$ and $(j,l)$ are identically
labeled edges, so selecting both candidates preserves one original edge.  This joint dependence
gives a quadratic objective.  Let $\aff$ be the sparse symmetric ACG adjacency, with incompatible
candidates as isolated vertices.  For a soft assignment $\soft\in[0,1]^{n\times m}$,
\begin{equation}
  J(\soft)=\vect{\soft}^{\top}\aff\vect{\soft}, \qquad
  \nabla J(\soft)=2\,\operatorname{unvec}(\aff\vect{\soft}),
  \label{eq:qap}
\end{equation}
where $\vect{\cdot}$ stacks rows and $\operatorname{unvec}$ inverts it.
A sparse scatter over ACG edges evaluates both quantities without materializing dense $\aff$.
For a compatible partial injection $\hard$, each selected ACG edge is counted in both
orientations, hence $J(\hard)=2|C(p)|$.
For a soft state, $\nabla J(\soft)_{ij}$ (the objective marginal of candidate $(i,j)$) sums
the support from other candidate matches that would preserve edges together with $i\mapsto j$.
It is computed exactly from the current state.

\subsection{Learning primal--dual matching updates}

\noindent\textbf{Matching scores and capacity prices.}
A bid $Z_{t,ij}$ is a real-valued score for matching source $i$ to target $j$,
and $z^\bot_{t,i}$ scores leaving $i$ unmatched.  Soft assignments $\soft_t$ and unmatched
probabilities $d_t$ are obtained by normalizing each source row:
\begin{equation}
  [\soft_{t,i1},\ldots,\soft_{t,im},d_{t,i}]
  =\operatorname{softmax}([Z_{t,i1},\ldots,Z_{t,im},z^\bot_{t,i}]).
  \label{eq:row-softmax}
\end{equation}
Each source distributes one unit between targets and ``unmatched,'' but several sources
may favor one target. Its excess demand is $q_{t,j}=\sum_iS_{t,ij}-1$.
A nonnegative price $\lambda_{t,j}$ penalizes it for every source.

The primal--dual interpretation comes from the capacity Lagrangian
$\mathcal F(S,\lambda)=J(S)-\langle\lambda,S^\top\one-\one\rangle$.
The assignments are primal variables and prices are multipliers for the column constraints.
Its assignment gradient contains $\nabla J(S)_{ij}-\lambda_j$, while descent in the price
variable increases prices under excess demand.  ENPDA retains these two opposing signals and
learns corrections and positive step sizes, with the precise fixed-state interpretation in
Appendix~\ref{app:dual-proof}.

\noindent\textbf{Initialization and shared policy.}
Initialization uses six structural features: node-label agreement, degree agreement, and multiset
similarities of neighbor labels, incident edge labels, and labeled paths of lengths two and three.
A fixed linear score supplies analytic logits, a negative incompatibility bias, and a fixed
unmatched logit. A neural encoder embeds each candidate match $(i,j)$ in 64 dimensions.
Two sparse ACG message-passing layers combine candidates that can jointly preserve edges.
Shared heads use these embeddings and their row, column, and global means to add logit
residuals, and prices start at zero. These symmetric operations respect independent input
relabelings. A separate encoder with the same architecture shares weights across recurrent
rounds. Both encoders are frozen at inference.
Output heads that map zero activations to a zero correction and a unit step multiplier
(Appendix~\ref{app:architecture}) make the untrained model exactly the analytic counterpart.

\noindent\textbf{One update round.}
We first compute the exact objective gradient and normalize it so that its scale is less
sensitive to graph size and density:
\begin{equation}
  \widetilde{G}_t=\frac{2\operatorname{unvec}(\aff\vect{\soft_t})}
  {\max\!\left\{\operatorname{RMS}_{(i,j)\in\mathcal C}(2\aff\vect{\soft_t}),10^{-6}\right\}},
  \qquad q_{t,j}=\sum_i S_{t,ij}-1.
  \label{eq:marginal-demand}
\end{equation}
RMS is the root mean square over compatible entries, and the $10^{-6}$ floor prevents division
by near-zero scales; when $\mathcal C$ is empty the RMS is undefined and the scale is set to one.
The dynamic encoder receives the six fixed features plus current assignment mass, normalized
marginal, price, excess demand, and row entropy.  It predicts a correction $c_{t,ij}\in[-1,1]$,
positive bid and price step multipliers $a_{t,ij},b_{t,j}\in[1/2,2]$, and an unmatched-logit
update $u_{t,i}\in[-1,1]$.  These predictions modify the following fixed update rules:
\begin{align}
  \lambda_{t+1,j}
    &=\Pi_{[0,20]}\!\left(\lambda_{t,j}+\tfrac12 b_{t,j}q_{t,j}\right),
    \label{eq:dual-update}\\
  Z_{t+1,ij}
    &=Z_{t,ij}+\tfrac12 a_{t,ij}
      \left(\widetilde G_{t,ij}+c_{t,ij}-\lambda_{t+1,j}\right)
      -0.1\,\mathbf 1\{(i,j)\notin\mathcal C\},
    \label{eq:primal-update}\\
  z^\bot_{t+1,i}&=z^\bot_{t,i}+\tfrac12 u_{t,i},
  \qquad (\soft_{t+1},d_{t+1})=\operatorname{RowSoftmax}(\bid_{t+1},z^\bot_{t+1}).
  \label{eq:dummy-update}
\end{align}
Here $\Pi_{[0,20]}$ clips prices and $\mathbf 1\{\cdot\}$ is the indicator function.
Prices rise with excess demand and fall toward zero under spare capacity. Bids then reward
edge preservation through $\widetilde G_t$, adjust preferences through $c_t$, and subtract
the new price. The unmatched logit lets a source withdraw mass from real targets.
The step multipliers are positive and enter separately: $a_{t,ij}$ rescales the bracketed
bid update $\widetilde G_{t,ij}+c_{t,ij}-\lambda_{t+1,j}$, while $b_{t,j}$ rescales the
price step $q_{t,j}$. The row softmax gives the next probabilities.

Unlike fixed linear-assignment scores, MCES marginals depend on the current matching,
so ENPDA recomputes them each round. The analytic counterpart removes initializer
residuals and sets $c=u=0$, $a=b=1$, retaining features, rounds, price updates, and projection
to isolate the effect of learning.

\begin{figure*}[t]
\centering
\resizebox{\textwidth}{!}{\begin{tikzpicture}[
  x=1mm,y=.75mm,
  arr/.style={-{Stealth[length=2.35mm,width=1.75mm]},draw=black!78,line width=.82pt,
    preaction={draw=white,line width=2.1pt}},
  box/.style={draw=enpdablue!55,rounded corners=1.4mm,fill=white,
    align=center,font=\small,inner sep=1.1mm},
  nbox/.style={box,fill=enpdablue!7,minimum width=15mm,minimum height=22mm},
  abox/.style={box,fill=black!1,minimum width=22mm,minimum height=22mm},
  molatom/.style={circle,draw=enpdablue!72!black,fill=white,line width=.7pt,
    minimum size=3.8mm,inner sep=0pt,font=\rmfamily\scriptsize},
  molcarbon/.style={molatom,fill=enpdablue!7},
  moloxygen/.style={molatom,fill=enpdateal!12,text=enpdateal!55!black},
  molnitrogen/.style={molatom,fill=enpdablue!13,text=enpdablue!70!black},
  molcenter/.style={molatom,draw=enpdared!80!black,fill=enpdared!78,
    text=white,font=\rmfamily\bfseries\scriptsize},
  molhydrogen/.style={molatom,fill=enpdared!8,text=enpdared!70!black},
  molbond/.style={draw=black!58,line width=.85pt,line cap=round},
  acgpair/.style={draw=enpdablue!65!black,fill=enpdablue!8,rounded corners=1.6mm,
    line width=.55pt,minimum width=5.4mm,minimum height=3.2mm,inner sep=.15mm,
    font=\rmfamily\fontsize{4.8}{5.2}\selectfont},
  acgcenter/.style={acgpair,draw=enpdared!80!black,fill=enpdared!76,text=white,
    font=\rmfamily\bfseries\fontsize{4.6}{5}\selectfont},
  acgoxygen/.style={acgpair,fill=enpdateal!14,text=enpdateal!55!black},
  acgnitrogen/.style={acgpair,fill=enpdablue!15,text=enpdablue!72!black},
]
  % Three quiet, low-saturation regions following NGA's visual grammar.
  \filldraw[draw=enpdablue!18,fill=enpdablue!4,rounded corners=4mm]
    (0,28) rectangle (77,61);
  \filldraw[draw=enpdablue!15,fill=black!2,rounded corners=4mm]
    (0,0) rectangle (77,26);
  \filldraw[draw=enpdateal!22,fill=enpdateal!4,rounded corners=5mm]
    (81,3) rectangle (216,58);

  \node[anchor=west,font=\normalsize\bfseries] at (2,58)
    {Association common graph construction};

  % Molecular input graph G_a.  The atom, bond, and stereobond grammar follows
  % the cleaner GSF-chi TikZ molecules used in the companion ChiDeK project.
  \coordinate (ac) at (12,44);
  \coordinate (al) at (5,44);
  \coordinate (ar) at (18,44);
  \coordinate (af) at (24,48);
  \coordinate (ao) at (13,52);
  \coordinate (ah) at (13,36.5);
  \draw[molbond] (al)--(ac)--(ar)--(af);
  \fill[black!55] (ac)--(11.9,50)--(14.1,50)--cycle;
  \foreach \yy/\half in {42.6/.20,41.3/.35,40.0/.50,38.7/.68,37.7/.86}
    \draw[black!55,line width=.5pt] ({13-\half},\yy)--({13+\half},\yy);
  \node[molcenter] at (ac) {$C^*$};
  \node[molnitrogen] at (al) {N};
  \node[molcarbon] at (ar) {C};
  \node[molcarbon] at (af) {C};
  \node[moloxygen] at (ao) {O};
  \node[molhydrogen] at (ah) {H};
  \node[font=\small] at (13,31.5) {$G_a$};

  % G_b shares the central labeled motif but has a longer oxygenated branch.
  \coordinate (bc) at (40,44);
  \coordinate (bl) at (33,44);
  \coordinate (br) at (46,44);
  \coordinate (bf) at (52,48);
  \coordinate (be) at (58,44);
  \coordinate (bo) at (41,52);
  \coordinate (bh) at (41,36.5);
  \draw[molbond] (bl)--(bc)--(br)--(bf)--(be);
  \fill[black!55] (bc)--(39.9,50)--(42.1,50)--cycle;
  \foreach \yy/\half in {42.6/.20,41.3/.35,40.0/.50,38.7/.68,37.7/.86}
    \draw[black!55,line width=.5pt] ({41-\half},\yy)--({41+\half},\yy);
  \node[molcenter] at (bc) {$C^*$};
  \node[molnitrogen] at (bl) {N};
  \node[molcarbon] at (br) {C};
  \node[molcarbon] at (bf) {C};
  \node[moloxygen] at (be) {O};
  \node[moloxygen] at (bo) {O};
  \node[molhydrogen] at (bh) {H};
  \node[font=\small] at (42,31.5) {$G_b$};

  % Sparse ACG: every node is a compatible atom pair; an edge means that the
  % corresponding bond exists in both molecules.
  \node[acgnitrogen] (acn) at (64.4,51) {$N,N$};
  \node[acgoxygen] (aco) at (73.3,51) {$O,O$};
  \node[acgcenter] (acs) at (68.7,44) {$C^*,C^*$};
  \node[acgpair] (acc) at (64.2,37.2) {$C,C$};
  \node[acgpair] (acd) at (73.3,37.2) {$C,C$};
  \draw[molbond] (acs)--(acn) (acs)--(aco) (acs)--(acc) (acc)--(acd);
  \node[font=\small] at (69.0,31.5) {ACG $\aff$};
  \draw[arr] (59.5,44)--(acs.west);

  % Initial representations: six feature planes -> initializer -> bids; prices start independently at zero.
  \node[anchor=west,font=\normalsize\bfseries] at (2,2.7)
    {Initial primal--dual state};
  \foreach \k/\fill in {0/black!4,1/enpdablue!7,2/enpdateal!7}
    \draw[fill=\fill,draw=enpdablue!48] (5+\k*1.4,10+\k*1.2) rectangle +(10,9);
  \node[font=\small] at (11,7.1) {$X\in\mathbb R^{n\times m\times6}$};
  \node[box,fill=enpdablue!7,minimum width=13mm,minimum height=11mm] (enc0) at (28.5,15)
    {$\Psi_{\theta,0}$};
  \draw[arr] (17.8,15)--(21.5,15);

  \foreach \r in {0,...,2}{
    \foreach \c in {0,...,3}{
      \pgfmathtruncatemacro{\shade}{5+7*mod(2*\r+\c,4)}
      \draw[fill=enpdablue!\shade,draw=enpdablue!48,line width=.45pt]
        (40+3.8*\c,9+3.8*\r) rectangle +(3.8,3.8);
    }
  }
  \node[font=\small] at (47.6,7.1) {bids $\bid_0$};
  \draw[arr] (35.4,15)--(38.8,15);

  \foreach \r in {0,...,3}
    \draw[fill=white,draw=enpdablue!48,line width=.45pt]
      (62,9+3*\r) rectangle +(3.6,3);
  \node[font=\small] at (63.8,7.1) {$\price_0=0$};

  % Repeated learned-policy / analytic-update blocks.
  \node[font=\normalsize\itshape] at (148,54.5) {$T$ shared-weight auction rounds};
  \draw[draw=black!60,line width=.7pt] (87,51.5)--(87,53)--(194,53)--(194,51.5);

  \node[nbox] (p1) at (93,31)
    {$\Psi_{\theta,\mathrm{dyn}}$\\[-.5mm]\scriptsize $c_t,a_t,b_t,u_t$};
  \node[abox] (u1) at (115,31)
    {dual update\\[-.4mm]\scriptsize Eq.~\eqref{eq:dual-update}\\[-.2mm]
     bid + softmax\\[-.4mm]\scriptsize Eqs.~\eqref{eq:primal-update}--\eqref{eq:dummy-update}};
  \draw[arr] (p1.east)--(u1.west);

  % Two explicit input ports: structure enters from above and the initial
  % primal--dual state from below.  The symmetric routing avoids diagonal
  % crossings at the first shared-weight block.
  \draw[arr] (77,44)--node[above,font=\small,pos=.18] {$\aff$}(82,44)
    --(82,31)--(p1.west);
  \draw[arr] (77,15)--node[below,font=\small,pos=.26] {$\bid_0,\price_0$}(93,15)--(p1.south);

  \draw[arr] (u1.east)--(130,31);
  \node[font=\Large] at (135,31) {$\cdots$};
  \node[font=\scriptsize] at (135,18) {$\bid_{t+1},\price_{t+1}$};
  \node[nbox] (pt) at (151,31)
    {$\Psi_{\theta,\mathrm{dyn}}$\\[-.5mm]\scriptsize $c_t,a_t,b_t,u_t$};
  \node[abox] (ut) at (176,31)
    {dual update\\[-.4mm]\scriptsize Eq.~\eqref{eq:dual-update}\\[-.2mm]
     bid + softmax\\[-.4mm]\scriptsize Eqs.~\eqref{eq:primal-update}--\eqref{eq:dummy-update}};
  \draw[arr] (139,31)--(pt.west);
  \draw[arr] (pt.east)--(ut.west);
  \node[font=\normalsize\bfseries] at (147,7.5)
    {shared neural policy + explicit primal--dual update};

  % Hard lower bound and repaired-price upper bound are separate, visible branches.
  \node[box,fill=enpdablue!5,minimum width=20mm,minimum height=10mm] (hung) at (204,40)
    {Hungarian $\rightarrow\hard$\\[-.7mm]\scriptsize exact $L(\hard)$};
  \node[box,fill=enpdateal!5,minimum width=20mm,minimum height=10mm] (cert) at (204,21)
    {repair $\price_T$\\[-.7mm]\scriptsize valid $U$};
  \draw[draw=black!78,line width=.82pt] (ut.east)--(190,31);
  \fill[black!70] (190,31) circle (.65pt);
  \draw[arr] (190,31)--(190,40)--(hung.west);
  \draw[arr] (190,31)--(190,21)--(cert.west);
\end{tikzpicture}}
\caption{ENPDA inference.  A frozen policy uses the ACG to initialize bids and guide four
primal--dual updates.  Hungarian projection returns a partial matching and its lower bound,
and repairing the final prices gives a valid upper bound.}
\label{fig:pipeline}
\end{figure*}
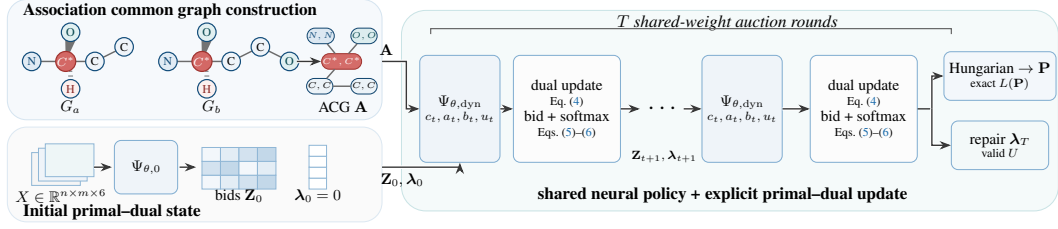

\noindent\textbf{From soft scores to a discrete matching.}
We use $T=4$ as a fixed inference budget, with longer trajectories evaluated separately.
After the updates, we append $n$ dummy columns to $\bid_T$, each containing $z_T^\bot$, and
apply one rectangular Hungarian projection.  Distinct dummy columns let every source remain
unmatched without competing for a shared slot.  Hungarian maximizes the sum of selected bids,
not the quadratic MCES objective.  We then count preserved edges.
The negative bias discourages incompatible
matches but does not forbid them, so restricting the output to endpoints of preserved compatible
edges gives a legal MCES witness with the same score (Theorem~\ref{thm:feasibility}).
Figure~\ref{fig:pipeline} and Appendix Algorithm~\ref{alg:enpda-core} summarize inference.

\subsection{Training the update policy}

\noindent\textbf{Where supervision comes from.}
Using only input graphs, a separately pretrained, frozen matcher proposes correspondences
that local search improves by exact edge count. These teacher maps are cached once;
they need not be optimal and use neither released correspondences nor optimum counts.
Appendix~\ref{app:architecture} specifies the reference matcher, its two continuous steps,
Hungarian rounding, and up to 30 refinement passes.

\noindent\textbf{Imitation relative to the analytic counterpart.}
For a teacher map $p^\star$, let $\mathcal I$ contain only source rows incident to its preserved
edges.  Supervising other rows would teach arbitrary matches that contribute no edge.
Define the negative log likelihood (NLL)
$\mathcal L_t=-|\mathcal I|^{-1}\sum_{i\in\mathcal I}\log S_{t,i,p^\star(i)}$, with
$\mathcal L_t=0$ for empty $\mathcal I$.  We supervise the whole trajectory and compare its
final state with the analytic trajectory on the same teacher rows:
\begin{align}
  \mathcal L_{\rm map}&=\mathcal L_T+\frac{0.25}{T}\sum_{t=1}^T\mathcal L_t,
    \label{eq:map-loss}\\
  \mathcal L_{\rm rel}&=[\mathcal L_T-\mathcal L^{\rm analytic}_T+0.01]_+ .
    \label{eq:relative-loss}
\end{align}
Intermediate supervision makes short trajectories useful, and the final state receives an additional
unit weight.  Here $\mathcal L^{\rm analytic}_T$ is the same NLL evaluated on the analytic
counterpart's final state, a fixed reference with no gradient.  When the learned NLL
fails to beat it by $0.01$, $\mathcal L_{\rm rel}$ adds imitation pressure.  This is a teacher-agreement margin.

\noindent\textbf{Edge preservation and competition.}
To retain a direct signal from the graph objective, define
$R_{\rm edge}=J(\soft_T)/(2M)$, where $M=\max\{\min(|E_a|,|E_b|),1\}$ sets the graph-size scale.
Soft states can violate target capacity, so this reward is a surrogate, not a feasible edge
count.  We penalize the worst overloaded target at each state:
$\mathcal L_{\rm cap}=(T+1)^{-1}\sum_{t=0}^T(\max_j[q_{t,j}]_+)^2$.
Finally, $\mathcal L_{\rm price}=m^{-1}\sum_j\lambda_{T,j}^2$ discourages unnecessarily large
prices.  The complete loss is
\begin{equation}
  \mathcal L=\mathcal L_{\rm map}+0.5\mathcal L_{\rm rel}
  -0.05R_{\rm edge}+0.1\mathcal L_{\rm cap}+10^{-4}\mathcal L_{\rm price}.
  \label{eq:training-loss}
\end{equation}
We backpropagate through the continuous rounds, while cached teachers and the Hungarian projection are
outside this gradient path.  Upper bounds are computed after inference and are not a
training-loss term.  We train on 1,343 pairs and select checkpoints on 172 validation pairs.  No labeled graph
identity is shared across training, validation, and test.  Checkpoint selection maximizes the
mean normalized hard-edge gain over the analytic counterpart at one, two, and four rounds.

\subsection{Core, Solver, and complexity}

ENPDA-Core runs one trajectory and one projection (Algorithm~\ref{alg:enpda-core}).
ENPDA-Solver runs four streams, each combining proposal generation with discrete search.  Three
streams use neural trajectories, one unperturbed and two with Gumbel-perturbed initial bids, and
one uses an analytic trajectory.  Each stream receives the same constructive, local-improvement,
simulated-annealing, and large-neighborhood budgets, and the largest preserved-edge count
determines the returned mapping.
The matched analytic Solver replaces all four trajectories with analytic ones under the same budgets.

For Core, with $h$ hidden channels and $T$ rounds, inference on an already constructed ACG costs
$O(T(nmh^2+|E_{\rm ACG}|h))$, followed by one $O(n^2m)$ rectangular Hungarian projection.
The $nm$ term includes incompatible grid entries.  This cost excludes feature/ACG construction
and Solver's discrete search.

\section{Theory}
\label{sec:theory}

Equivariance, feasibility, and the following bounds hold in exact arithmetic for every finite
network parameter value on the graphs of Section~\ref{sec:method}, regardless of training
quality. Full proofs appear in Appendix~\ref{app:theory}.

\begin{theorem}[Independent permutation equivariance]
\label{thm:equivariance}
For Core without random perturbations, let permutation matrices $\Pi\in\{0,1\}^{n\times n}$ and
$\Gamma\in\{0,1\}^{m\times m}$ independently reorder the
two input graphs.  At every round,
\[
  \soft_t(\Pi G_a,\Gamma G_b)=\Pi\soft_t(G_a,G_b)\Gamma^\top,
  \qquad \price_t(\Pi G_a,\Gamma G_b)=\Gamma\price_t(G_a,G_b).
\]
The unmatched state is row-equivariant, and the set of optimum augmented-Hungarian outputs is
equivariant under the same action.
\end{theorem}

Vertex numbering changes only continuous-state indexing. Hard equivariance concerns the
set of optimal projections. Index-based tie-breaking or floating-point reductions can
select different representatives. Appendix~\ref{app:numerics} tests numerical repeatability.

\begin{theorem}[Continuous and hard feasibility]
\label{thm:feasibility}
For every $t$ and source $i$, $S_{t,ij}\ge0$, $d_{t,i}\ge0$, and
$\sum_jS_{t,ij}+d_{t,i}=1$; prices remain in $[0,20]^m$.  The augmented Hungarian projection
returns a partial injective real-target assignment.  Removing incompatible or non-incident matched
vertices leaves the preserved-edge score unchanged and gives a compatible MCES witness.
\end{theorem}

Soft states may exceed column capacity, while hard injectivity comes from the Hungarian projection.
Finite incompatibility penalties do not themselves enforce label agreement, so the compatible
witness is obtained by the restriction in Theorem~\ref{thm:feasibility}.
Appendix~\ref{app:dual-proof} proves the fixed-state dual-step interpretation and a bound on
signed, step-weighted cumulative excess.

\noindent\textbf{A valid bound from neural prices.}
For node $i$, let $h^a_i(s)$ count edges of $G_a$ incident to $i$ with signature
$s=(\text{edge label},\text{opposite-node label})$, and let $h^b_j(s)$ count analogously in
$G_b$.  Define, on compatible candidate $(i,j)$,
\begin{equation}
  w_{ij}=\frac12\sum_s\min\{h^a_i(s),h^b_j(s)\}.
  \label{eq:incident-weight}
\end{equation}
Every preserved edge contributes one to the incident degree at each endpoint.  Therefore any legal
hard assignment $P$ satisfies $|C(P)|\le\sum_{ij}w_{ij}P_{ij}$.  Maximizing the right-hand side is
a partial linear assignment, whose dual has nonnegative row potentials $r_i$ and target potentials
$c_j$ with $r_i+c_j\ge w_{ij}$.
The relaxation counts locally compatible incident edges without requiring that all counted edges
coexist in one correspondence.  This explains both upper-bound validity and the slack that may
remain even after optimizing its assignment dual.

\begin{theorem}[Repaired-price anytime certificate]
\label{thm:price-certificate}
For any ENPDA price vector $\price\ge0$ and any scale $\alpha\ge0$, set
\begin{equation}
  c_j=\alpha\lambda_j,\qquad
  r_i=\max\!\left\{0,\max_{j:(i,j)\in\mathcal C}(w_{ij}-c_j)\right\}.
  \label{eq:price-repair}
\end{equation}
For a row with no compatible target, set $r_i=0$.  Then $(r,c)$ is dual feasible and
\[
  \operatorname{OPT}_{\rm MCES}\le U_{\rm price}(\alpha):=\sum_i r_i+\sum_jc_j.
\]
Taking the minimum over a nonempty finite set of scales, the two graph edge counts, an edge-signature
histogram bound, and the exact linear-assignment value remains a valid upper bound $U$.  If $L$ is
the edge count of ENPDA's returned hard mapping, $(L,U)$ is an anytime certificate and
$\lfloor U\rfloor=L$ proves global optimality.
\end{theorem}

For fixed incident weights and a chosen scale, price repair uses row maxima and summation.
It requires no additional optimization solve.  An optional Hungarian solve optimizes the same
linear relaxation and gives a bound at least as tight.  Learned prices are useful when a bound is
needed without this extra solve.  For tighter bounds we also use an independent sparse MILP
formulation with one variable per compatible candidate
and ACG edge.  Appendix~\ref{app:certificate-proof} derives this formulation and states the
numerical assumptions for solver-reported bounds.

\section{Experiments}
\label{sec:experiments}
\suppressfloats[t]

\subsection{Experimental setup}

\noindent\textbf{Data and metrics.}
We use released NGA MCES pairs, with AIDS and MCF-7 from TU
Datasets~\citep{morris2020tudataset} and MOLHIV from the Open Graph
Benchmark~\citep{hu2020open}, giving 100/91/100 native test pairs after predeclared exclusions.
The 1,343 training, 172 validation, and 291 test pairs share no labeled-graph identity
(Appendix~\ref{app:data}). Accuracy is the recovered percentage of the released reference
edge count, averaged over pairs. Additional metrics are Johnson-similarity MSE and runtime.

\noindent\textbf{Baselines and training.}
The analytic counterpart shares features, four rounds, and one projection. Additional controls
include Sinkhorn, no-price models, NGA, and classical search (Appendix~\ref{app:baselines}).
Three seeds each use fresh reference pretraining, teacher generation, and 12 student epochs;
validation alone selects checkpoints.

\noindent\textbf{Statistics and computation.}
Primary native comparisons use 20,000 graph-identity-cluster bootstrap draws with paired
seed resampling and pair-weighted means. The 95\% intervals are pointwise
(Appendix~\ref{app:data}).
GPU inference uses H100/CUDA 12.8, and classical search uses host CPUs.

\begin{table}[t]
\centering\small
\caption{MCES accuracy (\%) without complete-ring filtering.
(a) Local results; ENPDA averages three seeds. Nominal-60s settings are detailed in Table~\ref{tab:enpda-rascal-cap60}.
(b) NGA-study reports~\citep{ying2026neural} under their original settings, not paired with (a).
Bold highlights ENPDA's best values within each group.}
\label{tab:native-accuracy}
\label{tab:literature-accuracy}
\begin{minipage}[t]{0.49\linewidth}
\vspace{0pt}\centering
\textbf{(a) Our evaluation}\par\smallskip
\begin{tabular*}{\linewidth}{@{\extracolsep{\fill}}lrrr@{}}
\toprule Method & AIDS & MOLHIV & MCF-7\\
\midrule
\multicolumn{4}{@{}l}{\emph{No search}}\\
Analytic Core & 49.78 & 55.36 & 59.19\\
ENPDA-Core & \textbf{58.29} & \textbf{62.77} & \textbf{67.81}\\
\multicolumn{4}{@{}l}{\emph{Nominal 60s}}\\
RASCAL & 99.05 & 99.47 & 98.92\\
ENPDA & 98.82 & \textbf{99.50} & 98.49\\
\multicolumn{4}{@{}l}{\emph{Full solver runs}}\\
Analytic Solver & 99.57 & 99.68 & 99.31\\
NGA & 98.17 & 99.32 & 97.88\\
ENPDA-Solver & \textbf{99.67} & \textbf{99.81} & \textbf{99.40}\\
\bottomrule\end{tabular*}
\end{minipage}\hfill
\begin{minipage}[t]{0.47\linewidth}
\vspace{0pt}\centering
\textbf{(b) Reported by the NGA study}\par\smallskip
% Panel (b) of Table 1; shares its caption and table number.
\begin{tabular*}{\linewidth}{@{\extracolsep{\fill}}lrrr@{}}
\toprule
Method & AIDS & MOLHIV & MCF-7\\
\midrule
RASCAL & 90.67 & 88.74 & 90.26\\
FMCS & 60.63 & 67.99 & 71.54\\
McSplit & 61.32 & 72.74 & 69.23\\
GLSearch & 43.47 & 41.12 & 42.08\\
NGM & 33.34 & 52.21 & 42.38\\
GANN-GM & 49.76 & 72.18 & 63.47\\
GA & 70.99 & 71.09 & 72.92\\
Gurobi & 74.52 & 78.67 & 80.76\\
NGA & 98.64 & 99.20 & 97.94\\
\bottomrule
\end{tabular*}

\end{minipage}
\end{table}

\subsection{Do the learned updates improve proposals?}

\begin{figure}[t]
\centering
\includegraphics[width=0.60\linewidth]{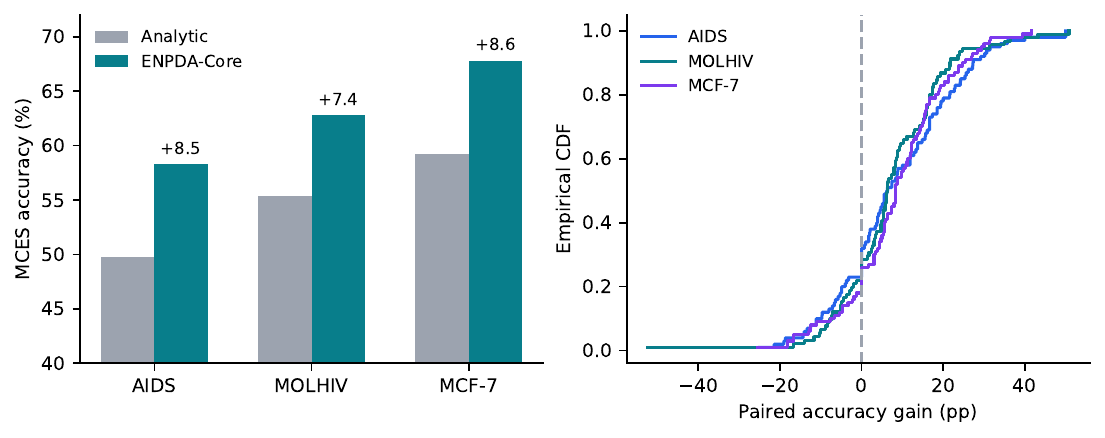}
\caption{Four-round Core accuracy (left) and per-pair learning gains averaged over three seeds
(right). More than half the pairs improve in each dataset; Table~\ref{tab:core-causality}
reports paired graph-cluster intervals.}
\label{fig:core-causality}
\vspace{-2.5mm}
\end{figure}

With the update structure and projection fixed, learning improves Core accuracy by
7.4--8.6 points on all three datasets, with all paired intervals excluding zero
(Figure~\ref{fig:core-causality}). Core takes 0.15--0.26 seconds per pair.
Deterministic reevaluation preserves these gains (Table~\ref{tab:deterministic-core}).

A train-once Sinkhorn control shares the split, teacher, width, and ACG encoder.
Core exceeds it on every dataset, with all paired intervals excluding zero
(Table~\ref{tab:train-once-sinkhorn}). Table~\ref{tab:components-transfer-main}a separates
the initializer, recurrent dynamics, and prices. Dynamics alone improve all three datasets,
while removing prices does not establish an accuracy loss, including after independent
retraining. The effect of prices on upper bounds is evaluated below. Eight rounds further improve
quality, while four rounds remain a deployment-budget choice (Table~\ref{tab:enpda-round-ablation}).

Frozen transfer uses 100 IMDB-BINARY pairs and 200 pairs each from PROTEINS and ENZYMES
with universal labels, 30\% edge deletion, and vertex permutation, giving known planted
optima. Without fine-tuning or search, Core gains 9.1--17.6 points
(Table~\ref{tab:components-transfer-main}b).
On 100 unedited IMDB pairs, conservative bounds support a positive learning effect even
though the optima of 52 pairs remain unresolved (Appendix~\ref{app:natural-imdb}).

All three assignment arms also return legal mappings on 25 natural D\&D pairs up to
592 nodes. The largest bin uses median 0.89 GiB incremental GPU memory.
These are scaling and feasible-bound results, not exact accuracies (Appendix~\ref{app:dd-scaling}).

Complete startup costs 1.84 allocated GPU-hours per checkpoint, including the teacher.
Core averages 0.19 seconds per query. Relative to three NGA fits per query,
the measured crossover is about 40 queries (Table~\ref{tab:amortized-cost}).

\begin{table}[t]
\centering\small
\caption{Four-round component accuracy (a) and frozen transfer to 500 planted-optimum pairs
(b), in percent. Settings, matched controls and paired intervals:
Tables~\ref{tab:enpda-component-ablation}, \ref{tab:enpda-nonmolecular-ood}
and~\ref{tab:enpda-protein-ood}.}
\label{tab:components-transfer-main}
\begin{minipage}[t]{0.51\linewidth}
\vspace{0pt}\centering
\textbf{(a) Molecular components}\par\smallskip
\begin{tabular*}{\linewidth}{@{\extracolsep{\fill}}lrrr@{}}
\toprule
Arm & AIDS & MOLHIV & MCF-7\\
\midrule
Analytic & 50.18 & 55.06 & 59.04\\
Initializer only & 54.15 & 57.66 & 59.94\\
Dynamics only & 55.84 & 60.10 & 64.88\\
Full without prices & 56.60 & 62.31 & 68.26\\
ENPDA (full) & \textbf{57.93} & \textbf{62.39} & 68.00\\
\bottomrule
\end{tabular*}
\end{minipage}\hfill
\begin{minipage}[t]{0.45\linewidth}
\vspace{0pt}\centering
\textbf{(b) Frozen topology transfer}\par\smallskip
\begin{tabular*}{\linewidth}{@{\extracolsep{\fill}}lrr@{}}
\toprule
Dataset & Analytic & ENPDA\\
\midrule
IMDB-BINARY & 68.06 & \textbf{77.14}\\
PROTEINS & 38.33 & \textbf{55.98}\\
ENZYMES & 37.98 & \textbf{55.53}\\
\bottomrule
\end{tabular*}
\end{minipage}
\end{table}

\subsection{Native end-to-end quality}

ENPDA-Solver reaches 99.4--99.8\% accuracy (Table~\ref{tab:native-accuracy}a),
exceeding paired NGA on all three datasets with intervals excluding zero
(Table~\ref{tab:native-controls}). It is faster than NGA on MOLHIV and MCF-7 and
slower on AIDS.
Table~\ref{tab:native-mse} reports similarity MSE. Figure~\ref{fig:case-studies} shows correspondences.

Under the same refinement budgets, the four-stream analytic Solver trails ENPDA-Solver by
0.09--0.13 accuracy points, with all paired intervals crossing zero
(Table~\ref{tab:native-controls}).

\begin{figure}[ht]
\begin{minipage}[t]{0.58\linewidth}
\vspace{0pt}\centering
\includegraphics[width=\linewidth]{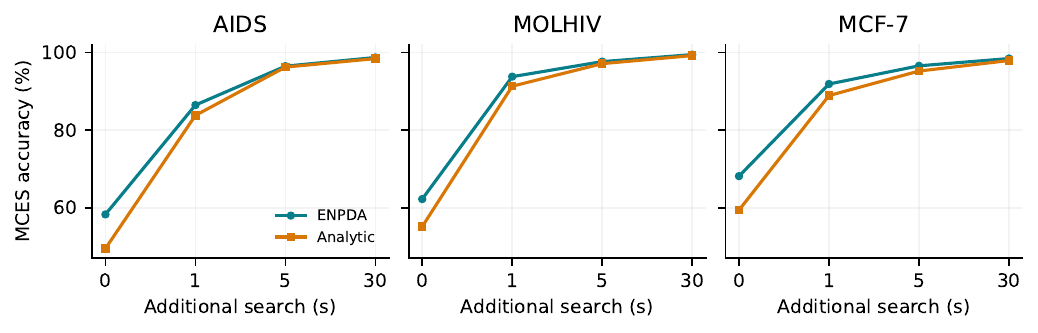}
\caption{Matched additional search after one four-round Core proposal and one projection.
Both arms use the same seeded search operators.}
\label{fig:search-frontier}
\end{minipage}\hfill
\begin{minipage}[t]{0.39\linewidth}
\vspace{0pt}\centering
\includegraphics[width=\linewidth]{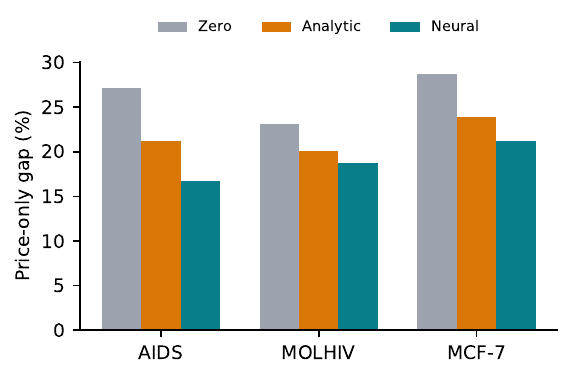}
\caption{Price-only gap (\%, $\downarrow$), fixed incumbent and repair.}
\label{fig:price-source-ablation}
\end{minipage}
\end{figure}

With one refinement stream applied identically to both arms, the initial learning gain remains
2.5--3 points after one second, with all intervals above zero
(Figure~\ref{fig:search-frontier}). At five and 30 seconds this holds only for MCF-7.
These are additional search budgets, so total time includes Core generation.

\noindent\textbf{Classical references.}
Table~\ref{tab:native-accuracy} separates our local evaluations from NGA-study reports.
The 60-second comparison uses method-specific stopping rules;
Table~\ref{tab:enpda-rascal-cap60} reports the realized runtimes.
Core-latency-capped FMCS and McSplit controls
have different objectives and are reported in Appendix~\ref{app:short-classical}.

\begin{figure}[t]
\centering
\includegraphics[width=0.82\linewidth]{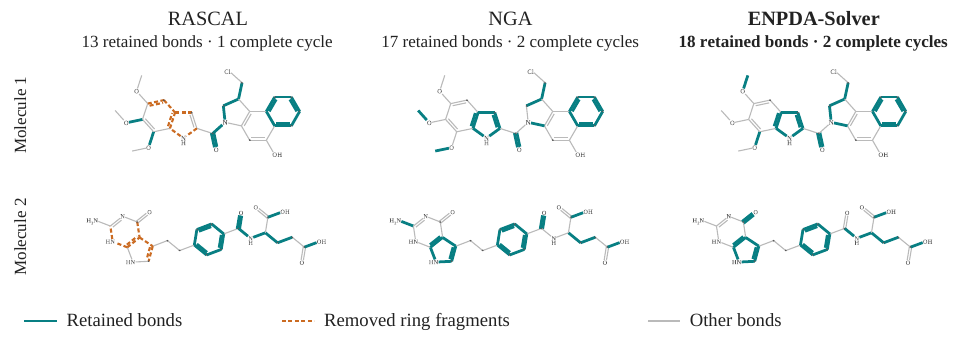}
\caption{MCF-7 pair 42 under the complete-cycle rule (Table~\ref{tab:aromatic-cycles}).
Teal bonds are retained; orange dashed bonds are compatible matches removed by the rule.}
\label{fig:aromatic-case-main}
\end{figure}

\noindent\textbf{Preserving complete aromatic rings.}
To exclude ring fragments, retained aromatic bonds must belong to complete cycles
mapped to complete target cycles. All methods receive a 60-second budget,
including projection and scoring (Appendix~\ref{app:aromatic-cycles}). ENPDA has higher
reference recovery on all datasets (Table~\ref{tab:aromatic-cycles}), taking 59.91 seconds
versus 12.61 for RASCAL and 17.78 for NGA. Table~\ref{tab:aromatic-cycles}
reports recovery relative to the original unrestricted references.
Figure~\ref{fig:aromatic-case-main} shows how the cycle rule removes fragmented aromatic matches.

\begin{table}[t]
\centering
\small
\caption{Complete-aromatic-cycle reference recovery (\%, $\uparrow$) relative to
Table~\ref{tab:native-accuracy}'s unrestricted references. The 60s budget includes
projection/scoring; neural seed 0, one NGA fit, RASCAL internal filter off
(Appendix~\ref{app:aromatic-cycles}).}
\label{tab:aromatic-cycles}
\begin{tabular}{lrccc}
\toprule
Dataset & Pairs & RASCAL & NGA & ENPDA-Solver\\
\midrule
AIDS & 100 & 86.85 & 78.46 & \textbf{89.73}\\
MOLHIV & 91 & 92.71 & 88.79 & \textbf{95.00}\\
MCF-7 & 100 & 86.27 & 82.10 & \textbf{89.04}\\
\bottomrule
\end{tabular}
\end{table}

\subsection{Hard tails and optimality bounds}
\label{sec:hard-certificates}

On the fixed 32-pair RASCAL tail unresolved at 900 seconds, full Solver gains 2.18
accuracy points (pooled 95\% interval $[0.03,4.46]$), taking 298 versus 903 seconds
(Table~\ref{tab:rascal-tail}).

At 1,800 seconds, the lifted relaxation establishes optimality for 47 of 75 fixed
Solver incumbents (Table~\ref{tab:long-certificate-enpda}). Repaired prices bound all 291
native pairs and establish 25 optima; structural and exact incident-assignment caps
raise this count to 60 (Table~\ref{tab:price-certificate-detail}).

With incumbent and repair fixed, neural prices reduce the mean gap
$100(U-L)/\max\{U,1\}$ versus analytic prices on all three datasets,
with all paired intervals excluding zero (Figure~\ref{fig:price-source-ablation}).
The exact incident-assignment cap dominates every repaired-price value.
Prices improve direct repair but not the strongest combined bound.

\subsection{Hard-negative retrieval}
\label{sec:retrieval}

On MOLHIV and MCF-7, over three graph-disjoint pools of 20 queries and 500 candidates per
dataset, ENPDA-Fast
reaches full-pool MAP of 0.957 (MOLHIV) and 0.926 (MCF-7). Its MAP lower bounds exceed
every baseline's upper bounds on full and hard-only pools; full-pool MRR and P@10 agree
(Figure~\ref{fig:enpda-retrieval}, Appendix~\ref{app:retrieval}). These bounds reflect unresolved
oracle labels, not sampling uncertainty. Appendix~\ref{app:retrieval} compares all six methods
on identical candidates and separates batch throughput from query latency.

\section{Conclusion}
\label{sec:conclusion}

We proposed ENPDA, an equivariant neural primal--dual method for maximum common edge
subgraph (MCES) search, coupling learned updates with feasible
projections and valid dual bounds. ENPDA improves fixed-depth MCES proposals,
short-budget refinement, and frozen transfer.  When matchings must keep aromatic rings intact, ENPDA retains more reference
bonds than all baselines.

\noindent\textbf{Future Work.}
Future work includes augmenting training pairs beyond the graph-isolated
AIDS graphs, scaling beyond the $nm$ grid, and establishing convergence
or approximation guarantees.

\ificlrfinal
\section{Acknowledgement}

This work is supported by SCION (Scientific Collaborative Innovation with Agentic Organizational Nexus) from Shanghai Innovation Institute.
\fi

\clearpage
\section*{AI use statement}
We used generative AI tools to assist with manuscript writing. Verification included implementation tests, mapping-feasibility and preserved-edge
checks, split-identity audits, deterministic reevaluation, and checks of reported values against saved outputs.
The authors take responsibility for the final text, mathematical claims, experimental results, and accompanying artifacts.

\ificlrrepro
\section*{Reproducibility statement}

Source code, configuration manifests, frozen per-run outputs, audit records,
and rendering scripts are available at
\url{https://anonymous.4open.science/r/ENPDA-MCES-3CEB}.
The artifact covers every training and evaluation pipeline in this paper:
dataset construction and graph-disjoint splits (Appendix~\ref{app:data}),
model architecture and training budgets (Appendix~\ref{app:architecture}),
baseline and comparison settings (Appendix~\ref{app:baselines}), and the
rescoring and audit coverage of Appendix~\ref{app:reproducibility}.  All
reported tables can be regenerated from the serialized manifests in the
repository.
\fi

\bibliographystyle{iclr2027_conference}
\bibliography{references}

\clearpage
\appendix
\section{Proofs and upper-bound derivations}
\label{app:theory}

\subsection{Notation table}
\label{app:notation}

Table~\ref{tab:notation} summarizes the mathematical notation.

\begin{table}[h]
\centering
\caption{Notation summary.}
\label{tab:notation}
\footnotesize
\setlength{\tabcolsep}{3.5pt}\renewcommand{\arraystretch}{1.08}
\begin{tabular}{@{}p{0.155\linewidth}@{}p{0.295\linewidth}@{\hspace{7pt}}p{0.17\linewidth}@{}p{0.315\linewidth}@{}}
\toprule
\multicolumn{2}{c}{Graphs and auction state} & \multicolumn{2}{c}{Network, training, and theory}\\
\midrule
$G_a=(V_a,E_a,\ell_a,\xi_a)$ & labeled source graph ($G_b$ analogously): vertices $V$, edges $E$, node labels $\ell$, edge labels $\xi$ &
$\theta$, $h$ & frozen network parameters; hidden width\\
$n$, $m$ & source and target vertex counts, $n=|V_a|\leq m=|V_b|$ &
$c_{t,ij}$, $a_{t,ij}$, $b_{t,j}$, $u_{t,i}$ & neural correction; primal and dual multipliers; unmatched update\\
$p$, $\bot$, $C(p)$ & partial injective map; unmatched marker; preserved-edge set &
$\Pi_{[a,b]}$, $[\cdot]_+$ & clipping projection; positive part\\
$\mathcal C$, $E_{\rm ACG}$ & label-compatible node pairs; association-common-graph edges &
$\one$, $\mathbf 1\{\cdot\}$, $\vect{\cdot}$ & all-ones vector; indicator function; row stacking\\
$\aff$ & sparse symmetric ACG adjacency on the $n{\times}m$ grid &
$p^\star$, $\mathcal I$, $\mathcal L_t$ & teacher mapping; active source rows; round-$t$ negative log likelihood\\
$\soft_t$, $\hard$ & soft assignment; hard partial-injection matrix &
$\mathcal L_{\rm map}$, $\mathcal L_{\rm rel}$, $\mathcal L$ & trajectory imitation, analytic-relative, and total losses\\
$\bid_t$, $z^\bot_{t,i}$, $d_{t,i}$ & real bids; unmatched logit; unmatched mass &
$R_{\rm edge}$, $\mathcal L_{\rm cap}$, $\mathcal L_{\rm price}$ & soft edge reward; capacity and price penalties\\
$\price_t$, $q_{t,j}$ & target prices; excess demand ($q_{t,j}>0$: over-demanded) &
$\Pi$, $\Gamma$ & independent input permutations (Section~\ref{sec:theory})\\
$\widetilde G_t$, $J(\cdot)$ & normalized ACG marginal; quadratic objective &
$\beta_{t,j}$, $B_t$, $\Lambda$, $o_{t,j}$ & deployed dual step; step matrix; price cap ($\Lambda{=}20$); clipped mass\\
$t$, $T$ & round index; number of rounds ($T{=}4$ deployed) &
$r_i$, $c_j$, $\alpha$ & repaired dual potentials; repair scale\\
$\mathcal F$, $\mathcal R$ & capacity Lagrangian; row-feasible soft domain &
$M$, $\mathcal L_T^{\rm analytic}$ & graph-size normalizer; fixed analytic teacher NLL\\
$h^a_i(s)$, $h^b_j(s)$, $w_{ij}$ & incident-signature counts and candidate weights &
$L$, $U$, $\mathrm{OPT}_{\rm MCES}$ & hard preserved-edge score; valid upper bound; optimal MCES value\\
\bottomrule
\end{tabular}
\end{table}

\subsection{The ACG quadratic equals twice the hard MCES score}

Write a candidate as $a=(i,j)$ and let $x_a=P_{ij}$.  The symmetric ACG adjacency obeys
$A_{ab}=1$ exactly when selecting both candidates $a=(i,j)$ and $b=(k,l)$ preserves the
corresponding labeled edges.  Therefore
\begin{align}
J(P)
 &=\sum_{a,b}A_{ab}x_ax_b
  =\sum_{\{a,b\}\in E_{\rm ACG}}(x_ax_b+x_bx_a) \nonumber\\
 &=2\sum_{\{a,b\}\in E_{\rm ACG}}x_ax_b
  =2|C(P)|. \label{eq:hard-qap-proof}
\end{align}
The partial-injection constraints prevent either endpoint of one source edge from being reused, so
each active ACG edge corresponds to one distinct preserved edge.  Equation~\eqref{eq:qap} follows
by replacing $P$ by $S$; differentiating the symmetric quadratic gives $2A\vect S$.

\subsection{Proof of permutation equivariance}

Let $K=\Pi\otimes\Gamma$ be the induced permutation of Cartesian candidates.  Relabeling the
input graphs preserves node and edge labels and hence transforms the candidate mask and ACG by
\begin{equation}
  \operatorname{mask}'=\Pi\operatorname{mask}\Gamma^\top,
  \qquad A'=KAK^\top. \label{eq:acg-action}
\end{equation}
The six fixed features compare only label equality, degrees, and multiset similarities of
neighbor labels, incident edge labels, and labeled simple paths of lengths two and three.
Consequently $X'=\Pi X\Gamma^\top$ on the first two tensor axes.

For one sparse message-passing layer, the transformed neighbor multiset of candidate $Ka$ is
exactly the image under $K$ of the neighbor multiset of $a$.  Shared pointwise maps and sum/mean
aggregation therefore give $H'=K H$.  Row means transform by $\Pi$, column means by $\Gamma$, and
the global mean is invariant.  The initial logits and unmatched logits hence satisfy
\begin{equation}
  Z'_0=\Pi Z_0\Gamma^\top,\qquad z_0^{\bot\prime}=\Pi z_0^\bot. \label{eq:init-action}
\end{equation}
Row softmax commutes with column reordering and independently with row reordering, establishing the
claim for $(S_0,d_0)$; $\lambda'_0=\Gamma\lambda_0=0$.

Assume the claim holds at round $t$.  Write $G_t=2\operatorname{unvec}(A\vect{S_t})$ for the
unnormalized marginal.  From Eq.~\eqref{eq:acg-action},
\begin{align}
 \vect{G'_t}
 &=2A'\vect{S'_t}
 =2KAK^\top K\vect{S_t}=K\vect{G_t},\\
 q'_t&=(S'_t)^\top\one-\one=\Gamma(S_t^\top\one-\one)=\Gamma q_t.
\end{align}
The compatible-entry RMS is invariant to reordering.  The dynamic encoder thus produces
$c'_t=\Pi c_t\Gamma^\top$, $a'_t=\Pi a_t\Gamma^\top$,
$b'_t=\Gamma b_t$, and $u'_t=\Pi u_t$.  Elementwise projection, multiplication, addition, and
price broadcasting preserve these actions in Eqs.~\eqref{eq:dual-update}--\eqref{eq:dummy-update}.
Induction proves Theorem~\ref{thm:equivariance} for every soft state.

For hard projection, write the augmented score matrix as
$\bar Z=[Z,z^\bot\one_n^\top]$ and let $\bar\Gamma=\operatorname{diag}(\Gamma,I_n)$.
Its relabeled version is $\bar Z'=\Pi\bar Z\bar\Gamma^\top$; dummy columns need no reordering
because they are identical.  If $Q$ is an augmented assignment, then
$Q'=\Pi Q\bar\Gamma^\top$ is feasible and has the same score.  This bijection proves
equivariance of the optimum set.  If the induced real-target partial map is unique, it also
transforms equivariantly.  An index-based tie rule need not have this property.

For stochastic proposal streams, the same induction holds with initial noise
$\varepsilon'=\Pi\varepsilon\Gamma^\top$.  Since the i.i.d.\ Gumbel law is invariant to these
permutations, the continuous states are equivariant in distribution.  This does not remove the
qualification about deterministic hard tie-breaking.

\subsection{Proof of feasibility}

Every component of a finite softmax is positive and its components sum to one.  Applying this fact
to Eq.~\eqref{eq:row-softmax} yields, for each row $i$,
\[
S_{t,ij}>0,\qquad d_{t,i}>0,\qquad \sum_jS_{t,ij}+d_{t,i}=1.
\]
Projection in Eq.~\eqref{eq:dual-update} directly gives $0\le\lambda_{t,j}\le20$.

For rounding, append $n$ dummy columns and copy source $i$'s unmatched logit to those columns.
The rectangular Hungarian problem assigns every source row to a distinct augmented column.  Real
columns are therefore used at most once; mapping a dummy selection to $\bot$ produces a partial
injection.  Incompatible selected candidates have no incident ACG edge.  Deleting them, and then
deleting any selected vertex not incident to a preserved ACG edge, cannot reduce $|C(P)|$ and
leaves a label-compatible MCES witness.  This proves Theorem~\ref{thm:feasibility}.

The equivariance induction and feasibility argument apply to every finite number of update
rounds.  Choosing four rounds in deployment is an experimental budget decision.

\subsection{Primal--dual interpretation and finite-horizon bound}
\label{app:dual-proof}

Let $\mathcal R=\{S\ge0:S\one\le\one\}$ be the row-feasible soft domain.  Relaxing column
capacities gives the Lagrangian
$\mathcal F(S,\lambda)=J(S)-\langle\lambda,S^\top\one-\one\rangle$ for $\lambda\ge0$.
For a fixed price vector, its assignment gradient is
$\nabla_S\mathcal F=\nabla J(S)-\one\lambda^\top$.  This explains the structural reward and
shared target penalty in Eq.~\eqref{eq:primal-update}.  The implementation normalizes the
objective gradient, adds learned corrections, and updates logits rather than $S$ directly.
Writing the real and unmatched logit increments as $\Delta Z_{ij}$ and $\Delta z_i^\bot$,
the resulting probabilities satisfy
\[
\begin{aligned}
 S_{t+1,ij}&=\frac{S_{t,ij}\exp(\Delta Z_{ij})}{D_i},\qquad
 d_{t+1,i}=\frac{d_{t,i}\exp(\Delta z_i^\bot)}{D_i},\\
 D_i&=\sum_jS_{t,ij}\exp(\Delta Z_{ij})+d_{t,i}\exp(\Delta z_i^\bot).
\end{aligned}
\]
Thus the primal update reweights each row multiplicatively while preserving its unit total,
including unmatched mass.

\begin{proposition}[Fixed-state projected price step]
\label{thm:dual-step}
Fix $S_t$ and its excess demand $q_t=S_t^\top\one-\one$.  With
$B_t=\tfrac12\operatorname{diag}(b_t)\succ0$, the price update is the diagonally scaled
projected gradient step on $\ell_t(\lambda)=-\langle\lambda,q_t\rangle$:
\[
 \lambda_{t+1}=\Pi_{[0,20]^m}
 \bigl(\lambda_t-B_t\nabla\ell_t(\lambda_t)\bigr).
\]
Each price is nondecreasing when its target is overloaded and nonincreasing when it is
underloaded, with equality permitted at a clipping boundary.
\end{proposition}

For $\ell_t(\lambda)=-\langle\lambda,q_t\rangle$,
$\nabla\ell_t(\lambda)=-q_t$.  With $B_t=\tfrac12\diag(b_t)$,
\[
\Pi_{[0,20]^m}\left(\lambda_t-B_t\nabla\ell_t(\lambda_t)\right)
=\Pi_{[0,20]^m}(\lambda_t+B_tq_t),
\]
which is Eq.~\eqref{eq:dual-update} and proves Proposition~\ref{thm:dual-step}.
The dual function of the continuous relaxation would be
$D(\lambda)=\max_{S\in\mathcal R}\mathcal F(S,\lambda)$.
An arbitrary network state $S_t$ need not attain this maximum, so $-q_t$ is not in general
a subgradient of $D$.  We claim a fixed-state Lagrangian update, not descent on $D$ or
monotonic improvement of $J$.  The price cap further restricts the deployed price dynamics;
upper-bound validity instead follows from the separate feasible repair below.

\begin{proposition}[Signed cumulative excess under clipping]
\label{thm:capacity}
Assume $\lambda_{0,j}=0$, positive deployed steps $\beta_{t,j}$, and
$\lambda_{t+1,j}=\Pi_{[0,\Lambda]}(\lambda_{t,j}+\beta_{t,j}q_{t,j})$ with $\Lambda=20$.
Let $o_{t,j}=[\lambda_{t,j}+\beta_{t,j}q_{t,j}-\Lambda]_+$ be upper clipping overflow.
Then for every horizon $T$ and target $j$,
\begin{equation}
 \left[\sum_{t=0}^{T-1}\beta_{t,j}q_{t,j}\right]_+
 \le\lambda_{T,j}+\sum_{t=0}^{T-1}o_{t,j}
 \le\Lambda+\sum_{t=0}^{T-1}o_{t,j}.
 \label{eq:capacity-bound}
\end{equation}
\end{proposition}

For Proposition~\ref{thm:capacity}, fix target $j$ and abbreviate
$x_t=\lambda_{t,j}+\beta_{t,j}q_{t,j}$.  Scalar clipping satisfies
\begin{equation}
  \Pi_{[0,\Lambda]}(x_t)\ge x_t-[x_t-\Lambda]_+ . \label{eq:clip-lower}
\end{equation}
Indeed, Eq.~\eqref{eq:clip-lower} is equality for $x_t\in[0,\Lambda]$ and $x_t>\Lambda$; for
$x_t<0$, its left side is zero and its right side is negative.  Hence
\[
\beta_{t,j}q_{t,j}\le\lambda_{t+1,j}-\lambda_{t,j}+o_{t,j}.
\]
Summing from $0$ to $T-1$, using $\lambda_{0,j}=0$, and noting that the right side is nonnegative
gives Eq.~\eqref{eq:capacity-bound}.  If no upper clipping occurs then all $o_{t,j}=0$.
More explicitly, if $v_{t,j}=[-x_t]_+$ is the lower clipping correction, telescoping gives
$\sum_t\beta_{t,j}q_{t,j}=\lambda_{T,j}+\sum_t o_{t,j}-\sum_t v_{t,j}$.
This is a bound on signed excess: under-demand can cancel over-demand.  It does not bound
$\sum_t\beta_{t,j}[q_{t,j}]_+$, ensure feasible columns at any particular round, or imply
convergence.  With varying steps it also does not directly bound unweighted cumulative demand.

\subsection{Repaired-price upper bound}
\label{app:price-proof}

For each source node $i$, let $\delta_i(P)$ be its degree in the preserved common-edge set under a
legal partial assignment $P$.  If $P_{ij}=1$, every one of these $\delta_i(P)$ edges must consume a
distinct incident edge at $i$ and at $j$ having the same edge label and the same opposite endpoint
label.  Thus
\begin{equation}
  \delta_i(P)\le\sum_s\min\{h_i^a(s),h_j^b(s)\}=2w_{ij}. \label{eq:degree-domination}
\end{equation}
The handshaking identity for the preserved subgraph and Eq.~\eqref{eq:degree-domination} give
\begin{equation}
  |C(P)|=\frac12\sum_i\delta_i(P)
  \le\sum_{ij}w_{ij}P_{ij}. \label{eq:linear-domination}
\end{equation}

Consider the partial linear assignment relaxation, with variables only on $\mathcal C$ and all
row and column sums restricted to compatible candidates:
\begin{align}
  \max_{x\ge0}\quad &\sum_{(i,j)\in\mathcal C}w_{ij}x_{ij}\\
  \text{s.t.}\quad &\sum_jx_{ij}\le1\quad(1\le i\le n),
  &\sum_ix_{ij}\le1\quad(1\le j\le m). \label{eq:linear-primal}
\end{align}
Its dual is
\begin{align}
  \min_{r,c\ge0}\quad &\sum_i r_i+\sum_jc_j\\
  \text{s.t.}\quad&r_i+c_j\ge w_{ij}\quad((i,j)\in\mathcal C). \label{eq:linear-dual}
\end{align}
For $c_j=\alpha\lambda_j\ge0$, Eq.~\eqref{eq:price-repair}, with $r_i=0$ for an empty row, is nonnegative and explicitly ensures
$r_i\ge w_{ij}-c_j$ for every candidate.  It is therefore feasible for
Eq.~\eqref{eq:linear-dual}.  Combining Eq.~\eqref{eq:linear-domination}, relaxation, and weak
duality yields
\begin{equation}
 |C(P)|\le\operatorname{OPT}_{\rm MCES}
 \le\operatorname{OPT}_{\rm assign}
 \le\sum_i r_i+\sum_jc_j=U_{\rm price}(\alpha). \label{eq:price-bound-chain}
\end{equation}
The feasible set in Eq.~\eqref{eq:linear-primal} is the bipartite partial-matching polytope and
has integral vertices.  Thus an exact maximum-weight partial assignment computes its LP optimum.
Zero-weight dummy columns allow unmatched sources.

Every chosen $\alpha$ is valid, so taking the minimum over any nonempty finite grid is valid,
including a grid chosen from the current graph and prices.  For the histogram cap, group edges
by the signature consisting of their edge label and the unordered pair of endpoint labels.
A preserved edge must match the same signature, giving the bound
$\sum_s\min\{H_a(s),H_b(s)\}$, where $H_a,H_b$ are the two signature histograms.
The edge-count caps and the exact optimum of Eq.~\eqref{eq:linear-primal} are also upper bounds;
their minimum remains valid.  If legal
incumbent $L$ and real upper bound $U$ obey $U<L+1$, integrality forces the optimum to equal $L$.
This proves Theorem~\ref{thm:price-certificate}.

The exact assignment value is no larger than any repaired-price dual value.  If it is included
in the minimum, the resulting bound is independent of the neural prices for fixed graphs.
The price-only ablation therefore evaluates the repair without this extra assignment solve.

\subsection{Sparse lifted formulation}
\label{app:certificate-proof}

The cheap bound linearizes incident capacity; an optional exact formulation represents ACG edges
directly.  For compatible candidate $a$ introduce $x_a\in\{0,1\}$; for ACG edge $e=\{a,b\}$
introduce $y_e\in[0,1]$:
\begin{align}
\max\quad &\sum_{e\in E_{\rm ACG}}y_e \label{eq:lifted-objective}\\
\text{s.t.}\quad
&\sum_{a:\,\operatorname{row}(a)=i}x_a\le1 &&\forall i,\\
&\sum_{a:\,\operatorname{col}(a)=j}x_a\le1 &&\forall j,\\
&y_{\{a,b\}}\le x_a,\quad y_{\{a,b\}}\le x_b &&\forall\{a,b\}\in E_{\rm ACG},\\
&y_{\{a,b\}}\ge x_a+x_b-1 &&\forall\{a,b\}\in E_{\rm ACG}. \label{eq:lifted-mccormick}
\end{align}
For binary $x$, the McCormick envelope makes $y_{\{a,b\}}=x_ax_b$, so every feasible integral
solution is a legal partial matching and Eq.~\eqref{eq:lifted-objective} is exactly its preserved
edge count.  Relaxing $x$ to $[0,1]$ gives an LP whose optimal value is an upper bound.
In exact branch-and-bound, every feasible integral incumbent is a lower bound and a valid global
bound from the active node relaxations is an upper bound.
Therefore
\[
 L_t\le\operatorname{OPT}_{\rm MCES}\le U_t,
 \qquad \operatorname{gap}_t=\frac{U_t-L_t}{\max\{U_t,1\}}
\]
is valid whenever these incumbent and bound conditions hold.  The formulation has
$|\mathcal C|+|E_{\rm ACG}|$ variables and $n+m+3|E_{\rm ACG}|$ structural inequalities,
in addition to variable bounds.

Our implementation uses double-precision HiGHS.  It records an LP value only after an optimal
solve and uses the MILP global dual bound for interrupted branch-and-bound runs.  Reported
values are inflated by $10^{-8}\max\{1,|U_{\rm raw}|\}$ and then rounded to the next larger
floating-point number.  Such outward adjustment preserves an already valid bound; it does not
independently verify the solver's floating-point certificates.  Solver-based optimality claims
therefore rely on the reported solver status and numerical tolerances, whereas the formulas
above establish the exact-arithmetic guarantee.

\section{Experimental details and additional results}
\label{app:experiments}

\subsection{Data scope, graph identity and statistics}
\label{app:data}

We retain the released integer node and edge labels.  AIDS and MCF-7 each contribute 100
native test pairs.  Seven of the 98 MOLHIV files contain empty released graph tensors;
reconstruction from stored SMILES changes their topology relative to the reference labels.
The predeclared keys 23, 46, 48, 54, 61, 64 and 76 are therefore excluded for every method,
leaving 91 pairs.  No exclusion uses model predictions.

The 291 native test pairs are fixed before graph-disjoint training inputs are constructed.
We exclude every test graph identity globally from reference training, student training and
validation, then separate training and validation graph identities.  Ordered tensor digests
cache repeats; four-iteration Weisfeiler--Lehman hashes prefilter exact isomorphism tests that
preserve node and edge labels.  Hash equality alone never establishes identity.
The resulting 1,343 training and 172 validation pairs share no graph identity with each other
or with test.  Every reference, teacher, full ENPDA, independently retrained no-price and
Sinkhorn model used in the native controls follows this split.
Table~\ref{tab:native-identity-audit} reports the retained scope, including the small AIDS
training pool.

\begin{table}[htbp]
\centering\small
\caption{Graph-disjoint native scope. Exact node/edge-label-preserving isomorphism checks are global across datasets; every pair of splits has zero shared graph identities. Test pairs are unchanged by the training split construction.}
\label{tab:native-identity-audit}
\setlength{\tabcolsep}{4pt}\renewcommand{\arraystretch}{1.04}
\begin{tabular}{lrrrr}
\toprule
Dataset & Train & Validation & Test & Test clusters\\
\midrule
AIDS & 19 & 11 & 100 & 21\\
MOLHIV & 628 & 76 & 91 & 90\\
MCF-7 & 696 & 85 & 100 & 82\\
\bottomrule
\end{tabular}
\end{table}

For native inference, connected components of the graph-identity incidence graph define
resampling units: pairs sharing either graph, directly or transitively, remain together.
There are 21, 90 and 82 components within AIDS, MOLHIV and MCF-7; the pooled native graph has
192 components, since one test graph identity appears in two datasets and merges their
components.  Each of 20,000 bootstrap replicates samples whole components with replacement
and independently samples experiment seeds with replacement, applying both draws jointly to the
compared methods.  The replicate mean divides summed pair effects by the number of sampled
pairs, preserving pair-weighted estimands.  We use percentile 95\% intervals, without a
simultaneous multiple-comparison guarantee.  Seed resampling is based on the three recorded
training runs. Bold highlights ENPDA's best values within each comparison, including ties,
without implying statistical significance. Other methods are not highlighted.

\subsection{ENPDA architecture and optimization}
\label{app:architecture}

Both ACG encoders use hidden width 64 and two sparse message-passing layers with SiLU activations
and LayerNorm.  The initializer encoder consumes six features.  The dynamic encoder consumes eleven:
the fixed six plus assignment mass, normalized ACG marginal, target price, target demand, and row
entropy.  Linear heads produce the correction, primal and dual multipliers, and unmatched update.
Row entropy includes unmatched mass and is divided by $\max\{\log(m+1),1\}$.
Every neural head is initialized at zero, so epoch zero is exactly the analytic counterpart.
In the feature order listed in Section~\ref{sec:method}, the analytic initializer uses weights
$(5,1.5,2,1,1.5,1)$ and bias $-1$, with an additional $-8$ for incompatible pairs; the unmatched
logit starts at $-1$.  The initial neural residuals are added to these scores.  Dynamic correction
and unmatched heads use $\tanh$, while step heads use $\exp(\log(2)\tanh(\cdot))$, giving a unit
multiplier at zero.  Before these nonlinearities, target step heads average over source rows and
unmatched heads over targets.  The analytic counterpart sets all neural residuals and corrections to zero and uses
unit multipliers throughout.

Training uses AdamW, learning rate $3\times10^{-4}$, weight decay $10^{-5}$, gradient accumulation
of eight, gradient clipping at 2, and 12 epochs.  In Eq.~\eqref{eq:training-loss}, final-state
negative log likelihood (NLL) receives an additional unit weight beyond the $0.25/T$ applied
to each round $t=1,\ldots,T$.  The remaining weights are 0.5 for the analytic-relative hinge,
0.05 for the soft edge reward, 0.1 for the capacity penalty, and $10^{-4}$ for the mean squared
final price.  The training seeds are 0, 1, and 2.
Probabilities in the implemented NLL are clipped below at $10^{-12}$ before taking logarithms.

Reference pretraining is performed afresh for each seed on the 1,343 training pairs only.
The eight-step mirror-assignment reference has hidden width 32 and is trained for 20 epochs
with learning rate $10^{-3}$ and gradient accumulation eight, maximizing the normalized ACG
soft objective.  Selection uses that training objective; neither validation graphs nor test
graphs enter reference optimization.  The student checkpoints are evaluated at epochs
0, 2, 4, 6, 8, 10 and 12, selecting mean normalized hard-edge gain over the fixed analytic
counterpart at one, two, and four rounds on the 172 validation pairs.
Full and no-price models select their own checkpoints (Table~\ref{tab:checkpoint-selection}).

For teacher generation we freeze the reference's learned six-feature initializer and positive
step schedule, replace the diagonal metric by the identity, and run two mirror updates with
partial Sinkhorn projection, without line search or the stationary safeguard.  One Hungarian
projection initializes up to 30 passes of steepest exact 2-exchange/reassignment, stopping when
no improving move exists.  One pseudo-map is cached for every training pair before student
optimization and is not refreshed with the student.  Only source rows incident to preserved
compatible edges contribute to mapping supervision; an empty active set has zero mapping loss.
Neither released correspondences nor optimum edge counts enter reference training, teacher
generation or student fitting.  The independently trained no-price arm fixes all prices to
zero throughout training and inference while keeping the same data, teacher, epoch budget and
other losses.  Inference-only removal instead evaluates the full checkpoint with zero prices.

\begin{table}[htbp]
\centering\small
\caption{Validation-selected epochs after matched 12-epoch training. Each arm selects its own checkpoint; no native test inputs enter selection.}
\label{tab:checkpoint-selection}
\setlength{\tabcolsep}{4pt}\renewcommand{\arraystretch}{1.04}
\begin{tabular}{lrr}
\toprule
Seed & Retrained no-price & ENPDA (full)\\
\midrule
0 & 10 & 4\\
1 & 12 & 12\\
2 & 12 & 6\\
\bottomrule
\end{tabular}
\end{table}

\begin{algorithm}[H]
\caption{ENPDA-Core inference}
\label{alg:enpda-core}
\begin{algorithmic}[1]
\REQUIRE Labeled graphs $G_a,G_b$; frozen shared parameters $\theta$; rounds $T$
\STATE Build the full node-pair feature grid, compatibility mask, and sparse ACG edges
\STATE $(\bid_0,z_0^\bot)\leftarrow$ analytic scores $+$ neural initializer; $\price_0\leftarrow0$
\STATE $(\soft_0,d_0)\leftarrow\operatorname{RowSoftmax}(\bid_0,z_0^\bot)$
\FOR{$t=0,\ldots,T-1$}
  \STATE compute exact ACG marginal $\widetilde G_t$ and demand $q_t=\soft_t^\top\one-\one$
  \STATE equivariant encoder predicts $(c_t,a_t,b_t,u_t)$
  \STATE update prices by Eq.~\eqref{eq:dual-update} and real and unmatched bids by Eqs.~\eqref{eq:primal-update}--\eqref{eq:dummy-update}
  \STATE $(\soft_{t+1},d_{t+1})\leftarrow\operatorname{RowSoftmax}(\bid_{t+1},z_{t+1}^\bot)$
\ENDFOR
\STATE $p\leftarrow\operatorname{Hungarian}([\bid_T,z_T^\bot\one_n^\top])$ using $n$ dummy columns
\STATE map every dummy-column selection in $p$ to $\bot$
\RETURN $p$, exact preserved-edge count, and target prices $\price_T$
\end{algorithmic}
\end{algorithm}

\subsection{Baseline details and timing conventions}
\label{app:baselines}

\paragraph{Classical and neural references.}
RASCAL reduces molecular MCES to maximum-clique search~\citep{raymond2002rascal}.
FMCS optimizes connected common bonds~\citep{dalke2013fmcs}, whereas the McSplit implementation
optimizes induced common vertices~\citep{mccreesh2017mcsplit}.  We score every returned map by
unrestricted labeled MCES.  GA uses graduated assignment~\citep{gold1996graduated}; Gurobi
optimizes an integer formulation~\citep{gurobi2026}.  GLSearch learns a subgraph-search
policy~\citep{bai2021glsearch}; NGM and GANN-GM learn association-graph matching
updates~\citep{wang2021ngm,wang2023unsupervised}.

NGA fits an optimizer separately to each query~\citep{ying2026neural}.  Our paired row uses the
same retained files, selects the highest-scoring map from three runs with a nominal 60-second
budget each, and charges the sum of all three fitting/inference times.  Its official code is
pinned to commit \texttt{e4a8f1f9ec9e31f79f3fbd648717dfbb9fe113fc}.
Cumulative amortization also counts all three fits per query.
The untrained analytic portfolio and per-pair NGA results need no graph-disjoint retraining.
Johnson MSE is recomputed from the verified hard-map similarity and the released reference,
using $(\hat s-s_{\rm ref})^2$ for every row.  Similarity-model literature rows use their
original settings: NeuroMatch~\citep{lou2020neuromatch}, SimGNN~\citep{bai2019simgnn},
GMN~\citep{li2019gmn}, XMCS~\citep{roy2022maximum} and INFMCS~\citep{lan2024interpretable}.

\paragraph{Train-once assignment controls.}
An independently trained affinity model shares ENPDA's graph-disjoint inputs, cached teacher,
hidden width 64 and two sparse ACG layers.  Validation selection uses normalized hard edges.
Sinkhorn applies 20 partial normalization iterations at temperature one and one Hungarian
projection.  Gumbel--Sinkhorn uses the same frozen affinity, ten independently seeded
standard-Gumbel perturbations, 20 Sinkhorn iterations and one projection per sample, then
selects the largest directly counted MCES.  Neither control performs post-projection search.
Table~\ref{tab:train-once-sinkhorn} compares these controls with primary Core.

\begin{table}[htbp]
\centering\small
\caption{Train-once controls using the new graph-disjoint split and teacher. Accuracy (\%), paired gain (pp) with 95\% CI, and mean measured time. Gumbel--Sinkhorn selects the best exact hard score from ten samples.}
\label{tab:train-once-sinkhorn}
\setlength{\tabcolsep}{4pt}\renewcommand{\arraystretch}{1.04}
\begin{tabular}{lrrr}
\toprule
Arm / contrast & AIDS & MOLHIV & MCF-7\\
\midrule
Sinkhorn & 45.82 & 47.53 & 45.60\\
Gumbel--Sinkhorn (10) & 39.51 & 44.83 & 46.30\\
ENPDA-Core & \textbf{58.29} & \textbf{62.77} & \textbf{67.81}\\
\midrule
Full $-$ Sinkhorn & +12.47 \confint{9.04}{15.42} & +15.24 \confint{12.05}{18.39} & +22.21 \confint{19.19}{25.42}\\
Full $-$ Gumbel--Sinkhorn & +18.77 \confint{15.95}{21.33} & +17.94 \confint{14.39}{21.68} & +21.52 \confint{18.01}{24.91}\\
\midrule
Sinkhorn time (s) & 0.253 & 0.151 & 0.169\\
Gumbel time (s) & 0.252 & 0.149 & 0.167\\
Core time (s) & \textbf{0.251} & \textbf{0.148} & \textbf{0.162}\\
\bottomrule
\end{tabular}
\end{table}

\paragraph{Measured intervals.}
Neural evaluation runs use NVIDIA H100 and CUDA 12.8, with the NVIDIA PyTorch 25.02 build
(PyTorch 2.7 development revision).  Native primary timing excludes disk and checkpoint loading
and includes ACG/feature construction, synchronized inference, projection, mapping transfer and
hard scoring.  Search measurements also include their declared discrete operators.
Worker-time totals, concurrent evaluation wall time and per-pair latency are reported separately.
The native and natural short-classical suites each run serially on their own GPU/CPU host with
one Torch/BLAS/search thread.  Their CPU models are Xeon Platinum 8558 and 8468, respectively;
we do not infer CPU/GPU speed ratios across suites.

\subsection{Primary causal controls and numerical repeatability}
\label{app:numerics}

Table~\ref{tab:core-causality} is the primary five-arm comparison.  In addition to removing
prices at inference, a separately trained no-price model controls for adaptation to their
absence.  The full-minus-retrained-no-price intervals all contain zero, so they do not establish
an independent accuracy contribution from prices or prove equivalence between the models.
The learning advantage over the analytic counterpart is positive in all three datasets.

\begin{table}[htbp]
\centering\small
\caption{Primary four-round, one-projection comparison, with no search. Accuracy is in percent; differences are percentage points with graph-cluster, paired-seed 95\% CIs. Price removal uses the full checkpoint; retrained no-price is a separately optimized model.}
\label{tab:core-causality}
\setlength{\tabcolsep}{4pt}\renewcommand{\arraystretch}{1.04}
\begin{tabular}{lrrr}
\toprule
Arm / contrast & AIDS & MOLHIV & MCF-7\\
\midrule
Analytic & 49.78 & 55.36 & 59.19\\
Retrained no-price & 58.81 & 62.45 & 67.73\\
Full, prices removed & 56.70 & 63.25 & 67.98\\
Analytic, no prices & 48.32 & 54.16 & 55.96\\
ENPDA-Core & 58.29 & 62.77 & 67.81\\
\midrule
$\Delta$: full $-$ analytic & +8.50 \confint{5.89}{12.31} & +7.41 \confint{4.32}{10.51} & +8.62 \confint{5.83}{11.43}\\
$\Delta$: full $-$ retrained no-price & -0.53 \confint{-2.81}{2.20} & +0.33 \confint{-1.95}{2.49} & +0.08 \confint{-1.71}{1.95}\\
$\Delta$: full $-$ prices removed & +1.59 \confint{-0.32}{3.97} & -0.47 \confint{-2.13}{1.25} & -0.16 \confint{-1.45}{1.05}\\
$\Delta$: analytic $-$ no prices & +1.46 \confint{-0.69}{3.51} & +1.21 \confint{-0.57}{2.99} & +3.23 \confint{1.36}{5.15}\\
Analytic seconds/pair & 0.247 & 0.144 & 0.159\\
Full seconds/pair & 0.251 & 0.148 & 0.162\\
\bottomrule
\end{tabular}
\end{table}

The mathematical equivariance statements concern exact arithmetic.  CUDA sparse reductions
can differ slightly across calls, and nearly tied Hungarian scores can turn small continuous
changes into different hard assignments.  Comparing the primary and independent deployment
measurements gives different edge counts on 396 of 873 learned seed--pair records and 227
analytic records, despite identical inputs and checkpoint hashes.  We therefore keep each
measurement with its own matched controls.

A repeatability diagnostic uses the first three lexicographically ordered native paths per
dataset with seed 0, comparing analytic and learned policies for eight repeated calls each.
Under default computation, four of nine learned cases and one of nine analytic cases have
multiple hard scores.  With \texttt{torch.use\_deterministic\_algorithms(True)} and
\texttt{CUBLAS\_WORKSPACE\_CONFIG=:4096:8}, all 18 cases have identical repeated continuous
states and hard mappings on this hardware.  This checks arithmetic repeatability, not
hard-map equivariance under tied projections or reproducibility across different hardware.

We then reevaluate all 291 pairs and all five primary arms across three seeds, at unchanged weights, with
these deterministic settings, yielding 4,365 independently rescored maps
(Table~\ref{tab:deterministic-core}).  This additional sensitivity analysis preserves both
main findings: positive learned-versus-analytic effects and unresolved full-versus-retrained-
no-price effects on all three datasets.

\begin{table}[htbp]
\centering\small
\caption{Deterministic arithmetic sensitivity: all 291 pairs and five arms across three seeds, with frozen full and no-price weights. Entries are accuracy (\%) or paired differences (pp) with 95\% CIs. These measurements supplement, rather than replace, the primary comparison.}
\label{tab:deterministic-core}
\setlength{\tabcolsep}{4pt}\renewcommand{\arraystretch}{1.04}
\begin{tabular}{lrrr}
\toprule
Arm / contrast & AIDS & MOLHIV & MCF-7\\
\midrule
Analytic & 48.53 & 55.31 & 59.61\\
Retrained no-price & 58.79 & 62.47 & 68.98\\
Full, prices removed & 57.86 & 62.06 & 68.39\\
Analytic, no prices & 47.18 & 54.57 & 55.89\\
Full & \textbf{59.15} & 62.15 & 67.96\\
\midrule
Full $-$ analytic & +10.62 \confint{7.06}{14.63} & +6.84 \confint{3.76}{9.83} & +8.35 \confint{5.34}{11.46}\\
Full $-$ retrained no-price & +0.36 \confint{-2.84}{3.21} & -0.31 \confint{-2.27}{1.50} & -1.01 \confint{-2.59}{0.45}\\
Full $-$ prices removed & +1.29 \confint{-0.97}{4.00} & +0.09 \confint{-1.48}{1.62} & -0.43 \confint{-2.05}{1.06}\\
Analytic $-$ no prices & +1.35 \confint{-0.48}{3.82} & +0.74 \confint{-0.89}{2.37} & +3.72 \confint{1.79}{5.64}\\
\bottomrule
\end{tabular}
\end{table}

\subsection{Components, rounds and deployment portfolios}

The matched component study intervenes on the initializer, recurrent corrections and prices
without retraining the full model.  Each arm retains four rounds and one projection.
Table~\ref{tab:enpda-component-ablation} gives all pairwise effects specified for this study;
individual gains do not imply that every component is necessary.

\begin{table}[htbp]
\centering\small
\caption{Separate matched component evaluation at four rounds and one projection. Accuracy (\%) and paired differences (pp), with 95\% CIs; these inference interventions do not retrain the full checkpoint.}
\label{tab:enpda-component-ablation}
\setlength{\tabcolsep}{4pt}\renewcommand{\arraystretch}{1.04}
\begin{tabular}{lrrr}
\toprule
Arm / contrast & AIDS & MOLHIV & MCF-7\\
\midrule
Analytic & 50.18 & 55.06 & 59.04\\
Initializer only & 54.15 & 57.66 & 59.94\\
Dynamics only & 55.84 & 60.10 & 64.88\\
Full without prices & 56.60 & 62.31 & 68.26\\
Full & \textbf{57.93} & \textbf{62.39} & 68.00\\
\midrule
Initializer $-$ analytic & +3.96 \confint{1.72}{6.47} & +2.60 \confint{0.03}{5.16} & +0.89 \confint{-1.93}{3.71}\\
Dynamics $-$ analytic & +5.66 \confint{3.07}{9.69} & +5.04 \confint{2.21}{7.78} & +5.83 \confint{2.83}{8.65}\\
No-price full $-$ analytic & +6.42 \confint{3.34}{10.34} & +7.25 \confint{4.23}{10.19} & +9.21 \confint{6.24}{12.11}\\
Full $-$ no-price full & +1.33 \confint{-0.94}{3.93} & +0.08 \confint{-1.70}{1.87} & -0.26 \confint{-2.43}{1.78}\\
Full $-$ initializer & +3.79 \confint{0.95}{7.81} & +4.72 \confint{2.21}{7.26} & +8.06 \confint{5.12}{11.09}\\
Full $-$ dynamics & +2.09 \confint{-0.45}{4.41} & +2.29 \confint{-0.36}{5.05} & +3.12 \confint{0.23}{6.00}\\
\bottomrule
\end{tabular}
\end{table}

The frozen model is also evaluated at one, two, four and eight rounds
(Table~\ref{tab:enpda-round-ablation}).  Eight rounds improve over four by 13.37, 6.68 and
9.67 points on AIDS, MOLHIV and MCF-7, respectively.  Four rounds are a fixed deployment
choice, not a convergence threshold.  The same-process timing study alternates warmed-up
four- and eight-round runs twice for every pair/checkpoint (Table~\ref{tab:enpda-round-latency}).
Feature construction and the initial encoder do not repeat with every dynamic update.

\begin{table}[htbp]
\centering\small
\caption{Round-budget sensitivity of the same frozen policy. Accuracy (\%) and learned-minus-analytic differences (pp), with 95\% CIs. Rows marked $8-4$ compare the learned eight- and four-round scores. Four rounds specify the deployment budget, not an optimal or converged trajectory.}
\label{tab:enpda-round-ablation}
\setlength{\tabcolsep}{4pt}\renewcommand{\arraystretch}{1.04}
\begin{tabular}{lrrrr}
\toprule
Dataset & Rounds & Difference [95\% CI] & Analytic & ENPDA\\
\midrule
AIDS & 1 & +1.83 \confint{-1.41}{5.89} & 41.33 & \textbf{43.16}\\
AIDS & 2 & +2.87 \confint{-0.39}{6.68} & 46.15 & \textbf{49.02}\\
AIDS & 4 & +9.05 \confint{5.83}{13.10} & 49.23 & \textbf{58.28}\\
AIDS & 8 & +10.98 \confint{8.47}{15.19} & 60.68 & \textbf{71.65}\\
AIDS & $8-4$ & +13.37 \confint{9.89}{17.73} & --- & ---\\
MOLHIV & 1 & +3.96 \confint{1.16}{6.79} & 43.40 & \textbf{47.36}\\
MOLHIV & 2 & +6.35 \confint{3.63}{9.07} & 48.51 & \textbf{54.86}\\
MOLHIV & 4 & +7.93 \confint{4.70}{11.08} & 54.84 & \textbf{62.77}\\
MOLHIV & 8 & +8.65 \confint{5.41}{11.82} & 60.80 & \textbf{69.45}\\
MOLHIV & $8-4$ & +6.68 \confint{4.05}{9.38} & --- & ---\\
MCF-7 & 1 & +3.75 \confint{0.70}{6.66} & 43.77 & \textbf{47.52}\\
MCF-7 & 2 & +7.34 \confint{4.28}{10.47} & 50.07 & \textbf{57.41}\\
MCF-7 & 4 & +8.80 \confint{5.77}{11.91} & 58.99 & \textbf{67.79}\\
MCF-7 & 8 & +13.27 \confint{9.85}{16.52} & 64.20 & \textbf{77.46}\\
MCF-7 & $8-4$ & +9.67 \confint{6.41}{12.75} & --- & ---\\
\bottomrule
\end{tabular}
\end{table}

\begin{table}[htbp]
\centering\small
\caption{Same-process H100 timing at four and eight rounds, after warmup; two repeats per pair/checkpoint. Mean seconds include ACG construction, synchronized trajectory, projection and hard scoring; disk loading is excluded.}
\label{tab:enpda-round-latency}
\setlength{\tabcolsep}{4pt}\renewcommand{\arraystretch}{1.04}
\begin{tabular}{lrrrrr}
\toprule
Dataset & Rounds & Build & Trajectory & Projection & Total\\
\midrule
AIDS & 4 & 0.0204 & 0.2349 & 0.0005 & 0.2558\\
AIDS & 8 & 0.0204 & 0.2396 & 0.0005 & 0.2606\\
MOLHIV & 4 & 0.0140 & 0.1370 & 0.0004 & 0.1515\\
MOLHIV & 8 & 0.0140 & 0.1420 & 0.0005 & 0.1565\\
MCF-7 & 4 & 0.0161 & 0.1503 & 0.0005 & 0.1669\\
MCF-7 & 8 & 0.0161 & 0.1551 & 0.0005 & 0.1716\\
\bottomrule
\end{tabular}
\end{table}

The complete Solver uses three learned streams, one with unperturbed and two with Gumbel-
perturbed initial bids, and one analytic trajectory.  Each stream receives 64 discrete restarts,
64 constructive restarts, up to 30 refinement passes, 2,500 annealing steps and 250
large-neighborhood steps; perturbation scale is 0.5.  The analytic Solver replaces all four
proposals with analytic trajectories under the same portfolio settings.  The highest verified
preserved-edge count determines the result.  This operator budget differs from the fixed
wall-clock caps used for RASCAL and the post-Core frontier.

\begin{table}[htbp]
\centering\small
\caption{Deployment measurements on the same 291 native pairs. Core calls here are independent measurements of the primary four-round policy. Solver and NGA use different optimization procedures; NGA time includes all three fits. Accuracy is percent; times are seconds. Differences include paired 95\% CIs.}
\label{tab:native-controls}
\setlength{\tabcolsep}{4pt}\renewcommand{\arraystretch}{1.04}
\begin{tabular}{lrrr}
\toprule
Arm / metric & AIDS & MOLHIV & MCF-7\\
\midrule
Analytic Core accuracy & 49.69 & 54.90 & 59.09\\
Analytic Core mean s & 0.25 & 0.15 & 0.16\\
Analytic Core P95 s & 0.42 & 0.27 & 0.36\\
ENPDA-Core accuracy & \textbf{57.97} & \textbf{61.80} & \textbf{67.87}\\
ENPDA-Core mean s & 0.26 & 0.16 & 0.17\\
ENPDA-Core P95 s & 0.44 & \textbf{0.27} & \textbf{0.36}\\
\midrule
Analytic Solver accuracy & 99.57 & 99.68 & 99.31\\
Analytic Solver mean s & 341.79 & 119.71 & 147.58\\
Analytic Solver P95 s & 593.91 & 244.08 & 357.28\\
NGA, three fits accuracy & 98.17 & 99.32 & 97.88\\
NGA, three fits mean s & 169.95 & 164.21 & 167.28\\
NGA, three fits P95 s & 173.02 & 168.84 & 170.88\\
ENPDA-Solver accuracy & \textbf{99.67} & \textbf{99.81} & \textbf{99.40}\\
ENPDA-Solver mean s & 200.83 & \textbf{115.55} & \textbf{131.04}\\
ENPDA-Solver P95 s & 434.91 & 236.36 & 330.55\\
\midrule
Solver $-$ analytic (pp) & +0.11 \confint{-0.05}{0.30} & +0.13 \confint{-0.03}{0.31} & +0.09 \confint{-0.15}{0.37}\\
Solver $-$ NGA (pp) & +1.51 \confint{1.12}{1.96} & +0.49 \confint{0.03}{1.03} & +1.52 \confint{0.78}{2.35}\\
Solver $-$ analytic (s) & -140.96 \confint{-148.77}{-129.25} & -4.16 \confint{-19.50}{20.23} & -16.54 \confint{-19.32}{-13.15}\\
Solver $-$ NGA (s) & +30.87 \confint{7.95}{63.85} & -48.66 \confint{-58.30}{-38.33} & -36.25 \confint{-60.98}{-8.00}\\
\bottomrule
\end{tabular}
\end{table}

\begin{table}[htbp]
\centering\small
\caption{Johnson-similarity MSE ($\times10^{-3}$). The first six rows retain literature settings. Paired NGA and both Solver rows use verified similarities and the same sanitized native reference files.}
\label{tab:native-mse}
\setlength{\tabcolsep}{4pt}\renewcommand{\arraystretch}{1.04}
\begin{tabular}{lrrr}
\toprule
Method & AIDS & MOLHIV & MCF-7\\
\midrule
NeuroMatch & 13.92 & 19.17 & 17.34\\
SimGNN & 15.42 & 14.38 & 20.29\\
GMN & 39.38 & 29.77 & 24.03\\
XMCS & 36.42 & 21.47 & 32.22\\
INFMCS & 25.18 & 16.39 & 31.12\\
NGA (literature) & 1.13 & 0.66 & 0.99\\
\midrule
NGA (paired) & 0.312 & 0.316 & 0.768\\
Analytic Solver & 0.114 & 0.175 & 0.200\\
ENPDA-Solver & \textbf{0.088} & \textbf{0.171} & 0.231\\
\bottomrule
\end{tabular}
\end{table}

\begin{table}[htbp]
\centering\small
\caption{Matched post-Core refinement. Budgets are additional search seconds, excluding Core. Accuracy (\%) and paired learned-minus-analytic differences (pp), with 95\% CIs.}
\label{tab:enpda-search-frontier}
\setlength{\tabcolsep}{4pt}\renewcommand{\arraystretch}{1.04}
\begin{tabular}{lrrrr}
\toprule
Dataset & Search s & Difference [95\% CI] & Analytic & ENPDA\\
\midrule
AIDS & 0 & +8.78 \confint{5.68}{13.37} & 49.53 & \textbf{58.32}\\
AIDS & 1 & +2.66 \confint{0.16}{6.59} & 83.79 & \textbf{86.44}\\
AIDS & 5 & +0.24 \confint{-0.47}{0.74} & 96.21 & \textbf{96.45}\\
AIDS & 30 & +0.29 \confint{-0.07}{0.64} & 98.39 & \textbf{98.69}\\
MOLHIV & 0 & +7.08 \confint{3.52}{10.60} & 55.18 & \textbf{62.26}\\
MOLHIV & 1 & +2.46 \confint{0.64}{4.66} & 91.29 & \textbf{93.76}\\
MOLHIV & 5 & +0.52 \confint{-0.04}{1.09} & 97.10 & \textbf{97.62}\\
MOLHIV & 30 & +0.25 \confint{-0.08}{0.62} & 99.17 & \textbf{99.42}\\
MCF-7 & 0 & +8.72 \confint{5.80}{11.68} & 59.43 & \textbf{68.16}\\
MCF-7 & 1 & +2.98 \confint{1.40}{4.64} & 88.88 & \textbf{91.86}\\
MCF-7 & 5 & +1.35 \confint{0.58}{2.13} & 95.19 & \textbf{96.54}\\
MCF-7 & 30 & +0.52 \confint{0.08}{0.99} & 97.90 & \textbf{98.41}\\
\bottomrule
\end{tabular}
\end{table}

Selected-source counts in Table~\ref{tab:enpda-stream-selection} include exact ties, which retain
the first candidate.  They describe where the returned solution was found.  The case figures retain the originally chosen
paths (AIDS 38, MOLHIV 96 and MCF-7 58) and redraw their saved mappings from the new seed-0
Solver.  Highlighted nodes are exactly the endpoints of preserved compatible edges.

\begin{table}[htbp]
\centering\small
\caption{Selected Solver stream over three seeds. Counts include ties, which retain the first candidate; these are selection frequencies, not causal attributions.}
\label{tab:enpda-stream-selection}
\setlength{\tabcolsep}{4pt}\renewcommand{\arraystretch}{1.04}
\begin{tabular}{lrrrr}
\toprule
Dataset & Analytic & Neural 0 & Neural 1 & Neural 2\\
\midrule
AIDS & 15 & 228 & 38 & 19\\
MOLHIV & 5 & 228 & 27 & 13\\
MCF-7 & 16 & 225 & 34 & 25\\
\bottomrule
\end{tabular}
\end{table}

\begin{figure}[htbp]
\centering
\includegraphics[width=0.85\linewidth]{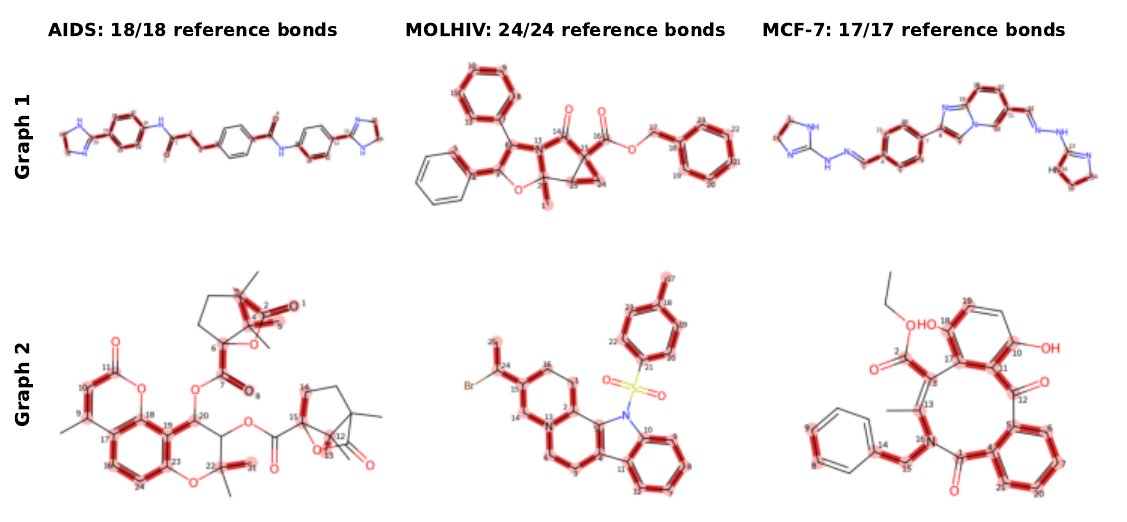}
\caption{Saved seed-0 ENPDA-Solver correspondences on the fixed example paths.  Red atoms and
bonds form the compatible common-edge witness; matching integers identify corresponding atoms.
All highlighted counts are recomputed from the saved maps.}
\label{fig:case-studies}
\end{figure}

\FloatBarrier
\subsection{Short-budget classical references}
\label{app:short-classical}

FMCS and McSplit run on the H100 host CPU using one search thread.  McSplit is pinned to commit
\texttt{a1f3e596ee8482ad332ff3de4166051338b5adaf}.  We retain its induced-vertex objective and
FMCS's connected-bond objective, and independently rescore their mappings as unrestricted MCES.
Timing includes conversion, features or search, projection when applicable and mapping validation,
excluding disk/checkpoint loading.  These independently timed Core calls retain their own
accuracy values and are not substituted into the primary comparison.

Fixed cooperative caps are 0.10, 0.16 and 0.60 seconds.  The matched cap for each pair is the
mean measured latency of its three Core seeds, fixed before evaluating the classical methods.
Every pair is retained, including early exits and overruns.  Table~\ref{tab:short-classical}
reports every method/cap, and Table~\ref{tab:short-classical-matched} reports paired gains,
P95 times, mean positive overruns and the fraction exceeding its cap by more than 1 ms.
The matched McSplit comparison supports a Core accuracy gain on MCF-7; all three FMCS
comparisons favor Core under their different objectives.

\begin{table}[htbp]
\centering\small
\caption{All native same-host short-budget controls. Entries are accuracy (\%) / mean milliseconds. Neural inference uses H100; classical search uses one host CPU thread. Search objectives differ, and all returned maps are scored by unrestricted MCES.}
\label{tab:short-classical}
\setlength{\tabcolsep}{4pt}\renewcommand{\arraystretch}{1.04}
\begin{tabular}{lrrr}
\toprule
Method & AIDS & MOLHIV & MCF-7\\
\midrule
Analytic-Core & 49.97 / 252.8 & 55.46 / 148.8 & 58.86 / 164.0\\
FMCS-0.10s & 44.48 / 22.7 & 46.62 / 7.5 & 47.19 / 15.1\\
FMCS-0.16s & 44.54 / 28.3 & 46.62 / 7.5 & 47.22 / 17.4\\
FMCS-0.60s & 44.56 / 52.9 & 46.62 / 7.6 & 47.22 / 35.4\\
FMCS-matched & 44.54 / 38.4 & 46.62 / 7.3 & 47.22 / 27.4\\
McSplit-0.10s & 56.32 / 93.4 & 63.32 / 88.8 & 60.09 / 90.5\\
McSplit-0.16s & 56.44 / 148.6 & 63.71 / 137.7 & 61.23 / 141.5\\
McSplit-0.60s & 57.67 / 541.0 & 65.96 / 482.2 & 62.97 / 482.2\\
McSplit-matched & 56.64 / 238.5 & 63.55 / 139.4 & 60.81 / 155.0\\
ENPDA-Core & 57.64 / 257.0 & 62.04 / 153.2 & \textbf{68.05} / 168.4\\
\bottomrule
\end{tabular}
\end{table}

\begin{table}[htbp]
\centering\small
\caption{Per-pair Core-latency caps. Gain is Core minus the classical reference (pp) with 95\% CI. P95 time and mean positive cap overrun are milliseconds; final column is the fraction exceeding cap by more than 1 ms.}
\label{tab:short-classical-matched}
\setlength{\tabcolsep}{4pt}\renewcommand{\arraystretch}{1.04}
\begin{tabular}{lrrrrr}
\toprule
Dataset & Reference & Gain [95\% CI] & P95 ms & Overrun ms & Over \%\\
\midrule
AIDS & FMCS & 13.10 \confint{8.59}{19.89} & 252.9 & 0.56 & 6.0\\
AIDS & McSplit & 1.00 \confint{-4.67}{6.00} & 418.1 & 0.34 & 1.0\\
MOLHIV & FMCS & 15.42 \confint{10.42}{20.06} & 17.0 & 0.00 & 0.0\\
MOLHIV & McSplit & -1.51 \confint{-7.09}{4.04} & 269.0 & 0.22 & 0.0\\
MCF-7 & FMCS & 20.82 \confint{16.54}{24.79} & 77.2 & 0.14 & 4.0\\
MCF-7 & McSplit & 7.24 \confint{3.30}{11.18} & 348.9 & 0.23 & 0.0\\
\bottomrule
\end{tabular}
\end{table}

\subsection{Non-molecular transfer with planted optima}
\label{app:nonmolecular-ood}

We select 100 IMDB-BINARY sources and 200 each from PROTEINS and ENZYMES, with 24--64 nodes.
Selection uses distinct source indices and fixed seeds before predictions.  Universal node and
edge labels remove molecular semantics.  A target retains
$\max\{1,\operatorname{round}(0.7|E|)\}$ randomly selected source edges and independently
permutes its vertices.  The planted correspondence preserves every target edge, meeting the
trivial target-edge upper bound and giving an exact optimum.  All three molecular checkpoints
remain frozen; both arms receive four rounds, one projection and no refinement.

Distinct indices can represent isomorphic topologies.  Exact-identity clustering gives 91,
200 and 197 pair-components for IMDB-BINARY, PROTEINS and ENZYMES; statistical intervals use
these clusters and paired seeds.  The protein source-selection seed is 20260916; the eligible
source pools contain 450 and 420 graphs, respectively.  Manifests retain source indices,
kept edges, permutations and hashes.  The 500 pairs yield 3,000 two-arm/three-seed records.
These controlled edge-deletion tasks measure topology transfer, not accuracy on arbitrary
natural protein pairs.

\begin{table}[htbp]
\centering\small
\caption{Frozen molecular-to-topology transfer under controlled 30\% edge deletion and node permutation. Accuracy (\%) uses the known planted optimum; differences (pp) have exact-identity-cluster, paired-seed 95\% CIs.}
\label{tab:enpda-nonmolecular-ood}
\setlength{\tabcolsep}{4pt}\renewcommand{\arraystretch}{1.04}
\begin{tabular}{lrrrr}
\toprule
Dataset & Pairs & Difference [95\% CI] & Analytic & ENPDA\\
\midrule
IMDB-BINARY & 100 & +9.07 \confint{7.18}{11.07} & 68.06 & \textbf{77.14}\\
\bottomrule
\end{tabular}
\end{table}

\begin{table}[htbp]
\centering\small
\caption{Frozen molecular-to-topology transfer under controlled 30\% edge deletion and node permutation. Accuracy (\%) uses the known planted optimum; differences (pp) have exact-identity-cluster, paired-seed 95\% CIs.}
\label{tab:enpda-protein-ood}
\setlength{\tabcolsep}{4pt}\renewcommand{\arraystretch}{1.04}
\begin{tabular}{lrrrr}
\toprule
Dataset & Pairs & Difference [95\% CI] & Analytic & ENPDA\\
\midrule
PROTEINS & 200 & +17.65 \confint{16.36}{18.93} & 38.33 & \textbf{55.98}\\
ENZYMES & 200 & +17.55 \confint{16.36}{18.74} & 37.98 & \textbf{55.53}\\
\bottomrule
\end{tabular}
\end{table}

IMDB exact recovery increases from 13.0\% to 18.7\%, with mean analytic/learned Core times of
0.701/0.705 seconds.  The PROTEINS and ENZYMES Core times, including ACG construction, are
0.198/0.202 and 0.186/0.190 seconds, respectively.  Hardware and graph structure differ across
these measurements, so they are not hardware-normalized timing comparisons.

\subsection{Unedited natural IMDB-BINARY pairs}
\label{app:natural-imdb}

We select 50 pairs from each half-open node bin $[10,20)$ and $[20,40)$.  Within a bin,
indices are sorted by SHA-256 of seed 20260916, the bin endpoints and index; the first 100
indices are paired consecutively.  No index is reused, no edges are edited and selection uses
neither labels nor method outcomes.  Universal labels and the same frozen molecular checkpoints
are used.  Exact topology grouping yields 54 components, the largest containing 36 pairs.
Neural inference uses H100; McSplit uses its single-thread host CPU with the fixed and per-pair
latency caps.  FMCS is absent from this entire suite because its fractional-time callback
failed to terminate during preflight; all selected pairs and all native FMCS runs are retained.

An independent oracle combines sparse lifted integer programming, structural upper bounds and
exact complete-graph reductions.  Combining its bounds with all validated incumbents closes
48 optima and leaves 52 unresolved.  For pair $k$, write its optimum interval as $[L_k,U_k]$.
A returned score $e_k$ gives accuracy in $[100e_k/U_k,100e_k/L_k]$; all 100 pairs enter the
mean, including unresolved cases.  For paired score difference $\delta_k$, take the smaller
and larger of $100\delta_k/L_k$ and $100\delta_k/U_k$, also when $\delta_k<0$.
Bootstrap the 54 graph components and paired seeds; the 2.5th percentile of lower effects and
97.5th percentile of upper effects give a conservative 95\% CI.

\begin{table}[htbp]
\centering\small
\caption{All 100 natural IMDB pairs. Accuracy ranges reflect unresolved optima; CIs for Core minus the row method also resample the 54 exact-topology clusters and training seeds. Timing is H100 plus host CPU, in milliseconds.}
\label{tab:natural-imdb-accuracy}
\setlength{\tabcolsep}{4pt}\renewcommand{\arraystretch}{1.04}
\begin{tabular}{lrrrr}
\toprule
Method & Accuracy range (\%) & Mean ms & P95 ms & Gain 95\% CI\\
\midrule
Analytic-Core & [83.94, 89.63] & 520.2 & 2663.3 & \confint{0.99}{4.66}\\
McSplit-0.10s & [77.79, 83.42] & 45.3 & 100.5 & \confint{1.03}{13.44}\\
McSplit-0.16s & [77.82, 83.47] & 69.8 & 160.6 & \confint{0.97}{13.38}\\
McSplit-0.60s & [78.23, 83.95] & 221.3 & 600.6 & \confint{0.20}{12.95}\\
McSplit-matched & [78.04, 83.73] & 198.3 & 390.8 & \confint{0.60}{13.12}\\
ENPDA-Core & \textbf{[86.28, 92.18]} & 526.9 & 2800.7 & ---\\
\bottomrule
\end{tabular}
\end{table}

Core's mean gain over the analytic counterpart lies in $[2.29,2.60]$ points, with conservative
95\% CI $[0.99,4.66]$.  The optimum-induced accuracy range and bootstrap CI express different
uncertainties.

\subsection{Natural non-molecular scaling}
\label{app:dd-scaling}

We use 25 pairs of independently observed D\&D protein graphs, five pairs per predeclared
size bin from roughly 50 to 500 nodes.  There is no planted correspondence or edit-derived
target, and exact MCES values are unknown.  We retain released discrete node labels and one
universal edge label.  Seed 0 is used for ENPDA and the train-once controls.  The legal edge
count divided by $U_0=\min(|E_1|,|E_2|)$ is a lower-bound diagnostic rather than exact accuracy.
All 75 method--pair mappings are independently validated.

\begin{table}[htbp]
\centering\small
\caption{Natural D\&D scaling, five pairs per size bin and seed 0. Entries are medians. Forward time excludes projection; ACG build is separate. $L/U_0$ is a lower-bound diagnostic, not exact MCES accuracy.}
\label{tab:enpda-dd-scaling}
\setlength{\tabcolsep}{4pt}\renewcommand{\arraystretch}{1.04}
\begin{tabular}{lrrrrrr}
\toprule
Nodes & Grid (k) & Compat. (k) & Build s & Forward s & Peak GiB & $L/U_0$\\
\midrule
63.5 & 4.0 & 0.2 & 0.19 & 0.88 & 0.045 & 0.171\\
111.0 & 12.3 & 0.6 & 0.58 & 2.58 & 0.073 & 0.183\\
188.5 & 35.1 & 2.3 & 2.15 & 8.49 & 0.152 & 0.215\\
310.5 & 95.8 & 5.2 & 5.71 & 25.15 & 0.360 & 0.211\\
501.0 & 249.0 & 13.9 & 14.59 & 62.48 & 0.886 & 0.242\\
\bottomrule
\end{tabular}
\end{table}

\begin{figure}[htbp]
\centering
\includegraphics[width=0.93\linewidth]{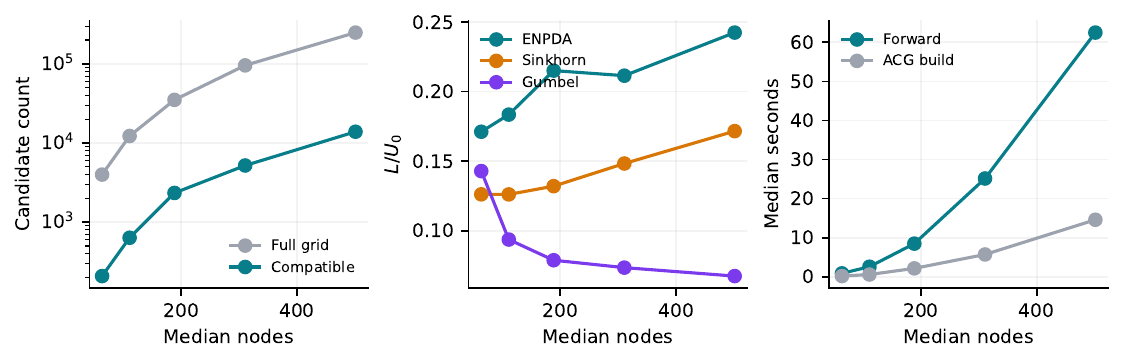}
\caption{Natural D\&D scaling.  Candidate sparsity, feasible lower-bound quality and measured
build/forward times are reported separately.  Each point is a five-pair median, with seed 0.}
\label{fig:enpda-dd-scaling}
\end{figure}

ENPDA has a higher edge count on all 25 pairs than Sinkhorn, and on 24 pairs than ten-sample
Gumbel--Sinkhorn, with one tie.  These are one-seed descriptive comparisons, plotted in
Figure~\ref{fig:enpda-dd-scaling}.
At the largest bin, median nodes are 501, full grid size is 249k, and 13.9k compatible
candidates induce 10.8k ACG edges.  Median ACG build time is 14.59 seconds; ENPDA forward
is 62.48 seconds, and projection plus hard scoring is 5.75 ms, with 0.89 GiB incremental GPU memory.
The full grid still incurs encoder cost even though ACG edges are sparse.

\FloatBarrier
\subsection{RASCAL caps and fixed tails}

Table~\ref{tab:enpda-rascal-cap60} reports nominal 60-second stopping settings on
all native pairs, retaining returned incumbents. RASCAL records lack incumbent
discovery times, and ENPDA's recorded incumbent times do not fully cover objective
construction and projection. The nominal limits therefore describe method-specific
stopping settings, not a verified common wall-clock cutoff.

\begin{table}[htbp]
\centering\small
\caption{Nominal 60-second per-pair stopping settings, retaining returned incumbents.
Realized runtimes are reported; timing details appear in the accompanying text.
Accuracy (\%), paired difference (pp) with 95\% CI, and mean realized seconds (RASCAL / ENPDA).}
\label{tab:enpda-rascal-cap60}
\setlength{\tabcolsep}{4pt}\renewcommand{\arraystretch}{1.04}
\begin{tabular}{lrrrr}
\toprule
Dataset & Difference [95\% CI] & Mean seconds & RASCAL & ENPDA\\
\midrule
AIDS & -0.23 \confint{-0.65}{0.34} & 16.52 / 60.27 & 99.05 & 98.82\\
MOLHIV & +0.03 \confint{-0.43}{0.55} & 9.94 / 60.15 & 99.47 & \textbf{99.50}\\
MCF-7 & -0.43 \confint{-0.95}{0.11} & 11.78 / 60.15 & 98.92 & 98.49\\
All & -0.22 \confint{-0.50}{0.09} & 12.83 / 60.19 & 99.14 & 98.92\\
\bottomrule
\end{tabular}
\end{table}

RASCAL defines the tail independently of ENPDA: 21/11/18 unresolved AIDS/MOLHIV/MCF-7 pairs
at 60 seconds, and 12/6/14 at 900 seconds.  Full Solver is compared with the retained RASCAL
incumbents on those fixed subsets, without a common consumed-compute claim.  In particular,
the 60-second tail does not cap ENPDA at 60 seconds.  The pooled 900-second-tail gain has
95\% interval $[0.03,4.46]$ points on the selected tail.  Tables~\ref{tab:rascal-tail} and~\ref{tab:tail-runtime} give every dataset
slice, including mean and P95 runtime.

\begin{table}[htbp]
\centering\small
\caption{Frozen RASCAL-timeout subsets. ENPDA uses its full Solver settings, not the RASCAL cap. Accuracy (\%) and paired differences (pp), with 95\% CIs; the subset definition is conditional on RASCAL.}
\label{tab:rascal-tail}
\setlength{\tabcolsep}{4pt}\renewcommand{\arraystretch}{1.04}
\begin{tabular}{lrrrrr}
\toprule
Cap s & Dataset & Pairs & Difference [95\% CI] & RASCAL & ENPDA\\
\midrule
60 & AIDS & 21 & +3.32 \confint{2.02}{4.33} & 95.49 & \textbf{98.82}\\
60 & MOLHIV & 11 & +4.24 \confint{1.88}{6.67} & 95.64 & \textbf{99.87}\\
60 & MCF-7 & 18 & +3.58 \confint{0.10}{7.04} & 94.00 & \textbf{97.58}\\
60 & All & 50 & +3.62 \confint{2.21}{5.06} & 94.99 & \textbf{98.60}\\
900 & AIDS & 12 & +1.34 \confint{-0.79}{3.17} & 96.83 & \textbf{98.16}\\
900 & MOLHIV & 6 & +2.24 \confint{0.00}{5.17} & 97.74 & \textbf{99.98}\\
900 & MCF-7 & 14 & +2.88 \confint{-1.59}{7.95} & 94.31 & \textbf{97.19}\\
900 & All & 32 & +2.18 \confint{0.03}{4.46} & 95.90 & \textbf{98.08}\\
\bottomrule
\end{tabular}
\end{table}

\begin{table}[htbp]
\centering\small
\caption{Realized runtime on the fixed timeout subsets (seconds). Each entry is mean / P95. The two algorithms need not consume their nominal cap.}
\label{tab:tail-runtime}
\setlength{\tabcolsep}{4pt}\renewcommand{\arraystretch}{1.04}
\begin{tabular}{lrrr}
\toprule
RASCAL cap & Dataset & RASCAL & ENPDA\\
\midrule
60 & AIDS & 63.22 / 71.45 & 339.66 / 510.65\\
60 & MOLHIV & 65.24 / 73.26 & 176.88 / 256.90\\
60 & MCF-7 & 61.01 / 61.03 & 281.55 / 443.14\\
60 & All & 62.87 / 71.72 & 282.93 / 497.95\\
900 & AIDS & 903.87 / 913.36 & \textbf{344.23} / \textbf{504.04}\\
900 & MOLHIV & 903.54 / 908.22 & \textbf{183.29} / \textbf{245.54}\\
900 & MCF-7 & 901.83 / 906.71 & \textbf{308.20} / \textbf{443.48}\\
900 & All & 902.92 / 910.07 & \textbf{298.29} / \textbf{497.14}\\
\bottomrule
\end{tabular}
\end{table}

\subsection{Complete aromatic-cycle output constraint}
\label{app:aromatic-cycles}

\paragraph{Motivation and scope.}
Atom and bond labels alone allow a common subgraph to contain only part of an
aromatic ring. For example, five matched bonds from a six-membered aromatic
cycle do not preserve that cycle. Requiring complete cycles is useful when
the intended similarity should retain intact aromatic motifs; RDKit also
provides a complete-aromatic-ring option for molecular MCES~\citep{rdkitRascalGuide}.
We evaluate this additional molecular constraint separately from unrestricted
MCES. It is an output-processing extension of the existing solvers: the ENPDA
checkpoint and internal search objective are unchanged. It does not require
all functional groups or an entire fused ring system to be preserved.

\paragraph{One explicit rule for all methods.}
We use all 291 valid public native test pairs (100 AIDS, 91 MOLHIV and 100
MCF-7), with no outcome-based exclusions. RDKit 2024.03.5 identifies the
symmetric-SSSR cycles whose bonds are all aromatic. Given an injective,
atom-label-compatible mapping, an aromatic bond is retained only if it belongs
to a complete source cycle whose entire image is a complete target cycle.
The output contains the union of these mapped cycles and compatible
nonaromatic bonds. A complete constituent cycle within a fused system is
allowed. The output includes both the atom map and an explicit bond subset.
For a native RASCAL candidate, projection may remove bonds but cannot add
bonds absent from its emitted witness. Every candidate receives this same
projection, and the best projected bond count is retained.

\paragraph{Solver settings and deadline.}
ENPDA uses the existing seed-0 checkpoint and four-stream Solver.
RASCAL uses similarity threshold zero, at most 100,000 candidate bond pairs,
and a 60-second internal timeout. We report both settings of its internal
\texttt{completeAromaticRings} filter, with the explicit complete-cycle output
constraint enabled in both. The filter-off arm is the main control: in the
pinned library, the internal ring-signature extraction can retain disconnected
degree-zero spectator atoms. A minimal counterexample, benzene versus benzene
plus a disconnected sodium atom, returns six bonds with the internal filter
off and zero with it on. Moreover, enabling the internal filter alone does not
verify the final full-cycle witness, so we apply the explicit rule to both arms.

NGA uses the pinned public MCS2 implementation, seed 0, and one fresh fit per
pair, with at most 200 epochs and ten assignment samples. We retain its released
hard-decoding schedule (epochs 100--200) and apply the same complete-cycle
projection to each decoded sample. Fresh model and optimizer initialization
and per-pair optimization are included in the 60-second budget. A run that
produces no timely hard answer retains the initial empty witness; a late decode
is never credited. NGA receives one H100 per query with no concurrent queries
on that GPU. This is a single-fit comparison, distinct from the three-fit
portfolio in the original unrestricted table.

An external watchdog terminates the worker process group at 60 seconds.
Molecular conversion, cycle detection, feature construction, inference or
instance optimization, search, hard decoding, projection and scoring count
toward this deadline. Imports, input disk loading, CUDA context initialization
and frozen checkpoint loading are excluded. A partial answer is credited if
its projected witness was completed by the deadline, even when search has not
finished; no later improvement is admitted. Each stored witness is independently
checked for injectivity, matching labels, complete-cycle coverage and its
admission timestamp. Supervisor times include process cleanup after the cutoff.

\paragraph{Results and interpretation.}
Table~\ref{tab:aromatic-cycles} reports original-reference recovery as
$100N^{-1}\sum_{k=1}^{N} b_k^{\rm ring}/b_k^{\rm ref}$, where $b_k^{\rm ring}$ is
the retained bond count after complete-cycle projection and $b_k^{\rm ref}$ is
the original unrestricted reference count used in Table~\ref{tab:native-accuracy}.
Ratios are computed per pair before averaging; this is not a ratio of mean counts.
The common denominator permits a percentage display for both settings but supplies
no constrained optimum.
Table~\ref{tab:aromatic-cycle-details} reports mean retained bonds and actual
supervisor time. Against RASCAL with its internal filter off and the same
complete-cycle output rule, ENPDA wins on 56 pairs, ties on 206 and loses on 29.
The pooled mean gain is 0.81 bonds, with a paired 95\% bootstrap interval of
$[0.52,1.13]$. These results cover the full test set for the seed-0 checkpoint.
All 291 NGA runs complete 200 epochs and return nonempty valid witnesses within
60 seconds; the maximum supervisor time is 32.15 seconds. Against NGA, ENPDA
wins on 176 pairs, ties on 110 and loses on five. The mean gain is 2.20 bonds,
with a paired 95\% bootstrap interval of $[1.83,2.59]$.
With the same complete-cycle output rule, ENPDA-Solver retains more bonds on average
than both baselines across all three datasets.
Mean times are 59.91 seconds for ENPDA, 12.61 for the RASCAL control and 17.78 for NGA.
Both use H100 GPUs where applicable and one CPU thread per worker. ENPDA and
RASCAL were evaluated in six shards, with all methods for each pair sharing its
shard hardware and four concurrent workers. CPU models across shards were
Xeon Platinum 8558, 8468 and 8462Y+; the software and CUDA 12.8 container were fixed.
The subsequent NGA runs use the same software on H100/Platinum 8462Y+,
with one query per GPU and one CPU thread per query.
The comparison establishes a retained-bond gain at a shared 60-second maximum budget.

\begin{table}[ht]
\centering
\small
\caption{Full-cycle comparison details. Each cell reports mean retained bonds /
mean supervisor seconds. All output witnesses satisfy the same complete-cycle rule,
including when RASCAL internal filtering is off. Difference rows give paired
95\% bootstrap intervals for mean bond gains (10,000 pair resamples).}
\label{tab:aromatic-cycle-details}
\begin{tabular}{lccc}
\toprule
Method & AIDS (100) & MOLHIV (91) & MCF-7 (100)\\
\midrule
RASCAL, internal filter off & 23.04 / 16.48 & 18.92 / 8.75 & 16.78 / 12.26\\
RASCAL, internal filter on & 17.63 / 10.06 & 18.16 / 6.57 & 15.49 / 10.72\\
NGA, one fit & 20.36 / 19.11 & 18.09 / 17.45 & 16.17 / 16.75\\
ENPDA-Solver & \textbf{24.12} / 60.36 & \textbf{19.38} / 59.56 & \textbf{17.63} / 59.79\\
\midrule
ENPDA minus RASCAL (off), 95\% CI & $[0.49, 1.78]$ & $[0.10, 0.90]$ & $[0.40, 1.33]$\\
ENPDA minus NGA, 95\% CI & $[2.90, 4.68]$ & $[0.98, 1.65]$ & $[1.07, 1.89]$\\
\bottomrule
\end{tabular}
\end{table}

\begin{figure}[p]
\centering
\includegraphics[width=\linewidth]{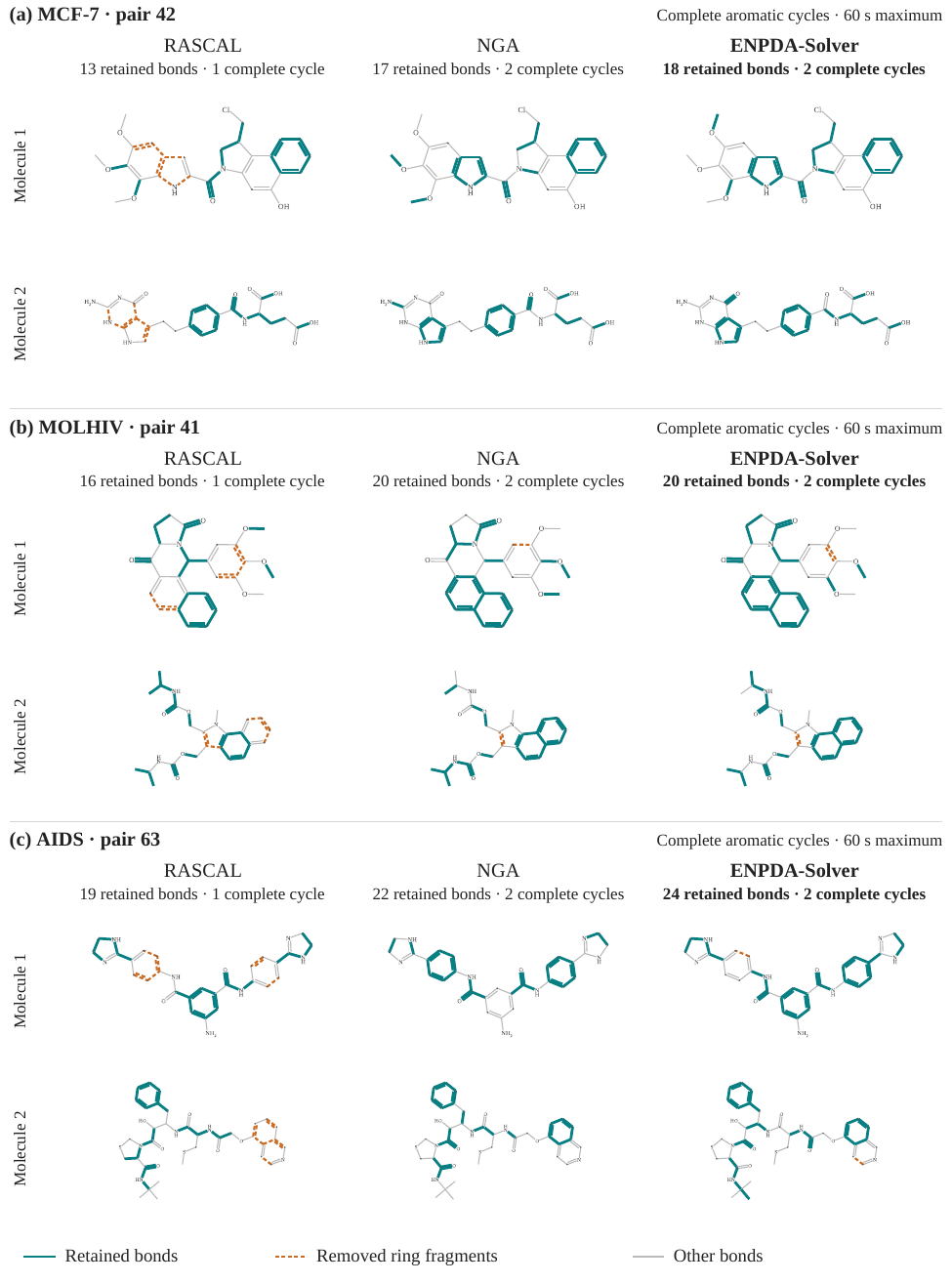}
\caption{Selected molecular examples from the complete-aromatic-cycle evaluation.
Teal bonds form the retained witness; dashed orange bonds are compatible under the
same candidate atom map but removed by the output rule; other bonds are gray.
Molecular coordinates are fixed across methods. The examples illustrate ring-fragment
removal and include an NGA--ENPDA tie; they are not a random sample of test performance.
All outputs satisfy the external 60-second cutoff. RASCAL uses its internal ring
filter off; ENPDA and NGA use seed 0, with one NGA fit.
Counts refer to retained bonds and complete aromatic cycles, not constrained optima.}
\label{fig:aromatic-cases}
\end{figure}

\paragraph{Molecular examples.}
Figure~\ref{fig:aromatic-cases} shows selected outputs from all three datasets,
including an NGA--ENPDA tie on MOLHIV pair 41. In MCF-7 pair 42, RASCAL's saved
candidate contains 19 compatible bonds, of which six are removed by the cycle
rule, leaving 13. NGA and ENPDA retain 17 and 18 bonds. Each highlighted bond is
checked against the saved atom map and explicit bond witness. These examples
illustrate the output constraint; the aggregate comparison uses all 291 pairs.

\subsection{Complete amortized compute}

Table~\ref{tab:amortized-cost} records each fresh startup stage on a single allocated H100.
Reference pretraining, teacher generation, student fitting/validation and setup together
average 6,622.34 seconds, or 1.8395 GPU-hours, for one deployable checkpoint.
The separately trained no-price control is an experimental ablation, not a deployment prerequisite.
The cost also excludes unrelated searches, hyperparameter exploration and the rest of the
paper's evaluation; it is a measured deployment-startup account rather than total research compute.

\begin{table}[htbp]
\centering\small
\caption{Complete startup cost of one deployable checkpoint. Per-seed values are synchronized seconds on one H100; the last column is mean allocated GPU-hours. Student timing includes validation. No-price ablation training and test evaluation are not required deployment stages.}
\label{tab:amortized-cost}
\setlength{\tabcolsep}{4pt}\renewcommand{\arraystretch}{1.04}
\begin{tabular}{lrrrr}
\toprule
Stage & Seed 0 s & Seed 1 s & Seed 2 s & Mean GPUh\\
\midrule
Reference pretraining & 5728.86 & 5297.39 & 5381.80 & 1.5193\\
Teacher generation & 466.28 & 466.05 & 441.52 & 0.1272\\
Student optimization & 696.99 & 712.68 & 673.19 & 0.1929\\
Setup overhead & 0.74 & 0.83 & 0.67 & 0.0002\\
Complete startup & 6892.87 & 6476.96 & 6497.19 & 1.8395\\
\bottomrule
\end{tabular}
\end{table}

For $Q$ subsequent queries, measured mean cost is
$C(Q)=1.83954+0.18816Q/3600$ allocated GPU-hours.
The paired NGA comparison counts all three fits, averaging 167.2 seconds per query, so
$\lceil6622.34/(167.2-0.18816)\rceil=40$ queries is the corresponding crossover.
At $Q=291$, costs are 1.85475 versus approximately 13.52 GPU-hours.

\subsection{Repaired-price and lifted bounds}
\label{app:price-certificate}

For lower bound $L$ and valid upper bound $U$, all gaps are
$100(U-L)/\max\{U,1\}$.  The price-source study fixes each seed's full Solver incumbent and
row repair, changing only zero, analytic-congestion or learned prices.  Across three seeds,
it contains 300, 273 and 300 seed--pair records.  The price-only upper bound uses scaled prices
and row maxima for incident weights.  Its validity is independent of learning.

\begin{table}[htbp]
\centering\small
\caption{Price-only ablation with identical Solver incumbents and row repair. Gaps are percentages, aggregated over three seeds; closed counts are seed--pair records with established optimality.}
\label{tab:price-source-ablation-detail}
\setlength{\tabcolsep}{4pt}\renewcommand{\arraystretch}{1.04}
\begin{tabular}{lrrrr}
\toprule
Dataset & Prices & Mean gap & Median gap & Closed\\
\midrule
AIDS & Zero & 27.14 & 23.26 & 20/300\\
AIDS & Analytic & 21.19 & 16.92 & 23/300\\
AIDS & Neural & \textbf{16.67} & \textbf{15.34} & \textbf{31/300}\\
MOLHIV & Zero & 23.11 & 23.08 & 3/273\\
MOLHIV & Analytic & 20.10 & 19.23 & 6/273\\
MOLHIV & Neural & \textbf{18.70} & \textbf{18.87} & \textbf{17/273}\\
MCF-7 & Zero & 28.70 & 28.16 & 15/300\\
MCF-7 & Analytic & 23.85 & 23.81 & 21/300\\
MCF-7 & Neural & \textbf{21.26} & \textbf{21.88} & \textbf{26/300}\\
\bottomrule
\end{tabular}
\end{table}

\begin{table}[htbp]
\centering\small
\caption{Paired price-only gap changes (pp), with graph-cluster and paired-seed 95\% CIs. Negative differences mean tighter upper bounds.}
\label{tab:price-source-contrasts}
\setlength{\tabcolsep}{4pt}\renewcommand{\arraystretch}{1.04}
\begin{tabular}{lrr}
\toprule
Dataset & Neural $-$ analytic & Neural $-$ zero\\
\midrule
AIDS & -4.51 \confint{-6.78}{-1.36} & -10.47 \confint{-14.64}{-5.92}\\
MOLHIV & -1.40 \confint{-2.43}{-0.51} & -4.41 \confint{-6.10}{-2.92}\\
MCF-7 & -2.59 \confint{-4.11}{-1.32} & -7.44 \confint{-10.14}{-5.19}\\
\bottomrule
\end{tabular}
\end{table}

Learned prices tighten this direct-repair alternative relative to analytic prices on all three
datasets.  The exact incident-assignment bound is independently recomputed and dominates every
repaired-price value.  Accordingly, price-source improvements do not further tighten the combined
bound.  For fixed incident weights, row repair is a simple maximum-and-sum computation;
timing only this repair excludes constructing the weights and solving an optional extra assignment.

\begin{table}[htbp]
\centering\small
\caption{Seed-0 upper bounds over all native pairs, using the full Solver incumbent. Price repair requires no further search. Combined additionally intersects structural and exact incident-assignment caps; gaps are median percentages.}
\label{tab:price-certificate-detail}
\setlength{\tabcolsep}{4pt}\renewcommand{\arraystretch}{1.04}
\begin{tabular}{lrrrrr}
\toprule
Dataset & Pairs & Price gap & Closed & Combined gap & Closed\\
\midrule
AIDS & 100 & 15.04 & 11 & \textbf{7.93} & \textbf{28}\\
MOLHIV & 91 & 18.28 & 5 & \textbf{10.34} & \textbf{10}\\
MCF-7 & 100 & 21.85 & 9 & \textbf{9.84} & \textbf{22}\\
\bottomrule
\end{tabular}
\end{table}

\begin{figure}[htbp]
\centering
\includegraphics[width=0.93\linewidth]{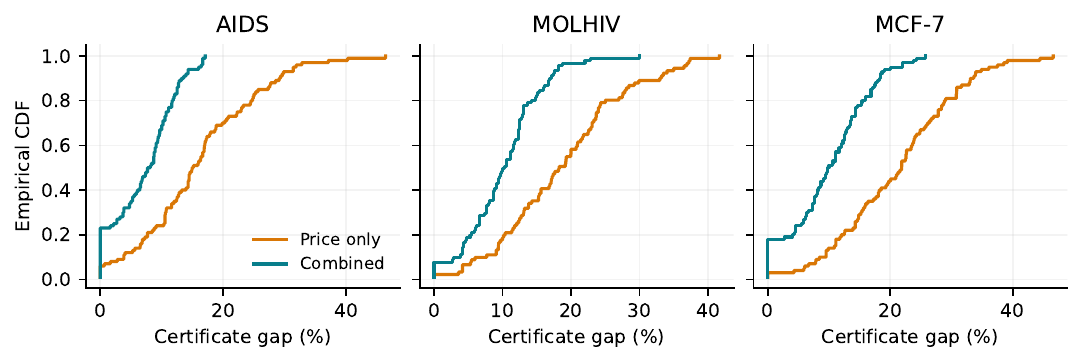}
\caption{Seed-0 optimality-gap bounds.  Price-only repairs learned prices; combined also
uses independent structural and exact incident-assignment caps.  Both use the same full Solver
lower bound.}
\label{fig:price-certificate}
\end{figure}

The seed-0 summary and Figure~\ref{fig:price-certificate} cover all 291 native pairs with the
full Solver incumbent: price-only closes 25, combined closes 60.  ``No additional search''
refers to bound construction and does not remove the search already used to obtain $L$.

The independent lifted subset contains 25 paths per dataset, selected from numeric file order
at zero-based positions $\operatorname{rint}(k(N-1)/24)$ for $k=0,\ldots,24$ and native scope
size $N$.  Rounding uses ties to even.  At each budget the bound is the minimum of valid
independent bounds available at that or smaller budgets.  We compare this bound with the
frozen seed-0 Solver score, without crediting stronger incumbents discovered by the bound
solver.  The 10/300/1,800-second envelopes establish optimality for 9/39/47 of 75 fixed incumbents.
The floating-point status/tolerance assumptions are stated in Appendix~\ref{app:certificate-proof}.

\begin{table}[htbp]
\centering\small
\caption{Independent lifted bounds compared with the frozen seed-0 Solver incumbent, on 25 fixed pairs per dataset. Bounds form an envelope over the current and smaller budgets. Gaps are percentages; newly discovered search incumbents receive no credit.}
\label{tab:long-certificate-enpda}
\setlength{\tabcolsep}{4pt}\renewcommand{\arraystretch}{1.04}
\begin{tabular}{lrrrrr}
\toprule
Budget s & Dataset & Pairs & Closed & Median gap & Mean gap\\
\midrule
10 & AIDS & 25 & 5 & 33.67 & 28.63\\
10 & MOLHIV & 25 & 1 & 30.56 & 31.75\\
10 & MCF-7 & 25 & 3 & 30.30 & 30.72\\
10 & All & 75 & 9 & 32.26 & 30.37\\
300 & AIDS & 25 & 14 & 0.00 & 13.81\\
300 & MOLHIV & 25 & 13 & 0.00 & 13.48\\
300 & MCF-7 & 25 & 12 & 5.00 & 16.59\\
300 & All & 75 & 39 & 0.00 & 14.63\\
1800 & AIDS & 25 & 17 & 0.00 & 10.62\\
1800 & MOLHIV & 25 & 14 & 0.00 & 11.17\\
1800 & MCF-7 & 25 & 16 & 0.00 & 12.04\\
1800 & All & 75 & 47 & 0.00 & 11.28\\
\bottomrule
\end{tabular}
\end{table}

\FloatBarrier
\subsection{Hard-negative retrieval}
\label{app:retrieval}

\begin{figure}[t]
\centering
\includegraphics[width=0.94\linewidth]{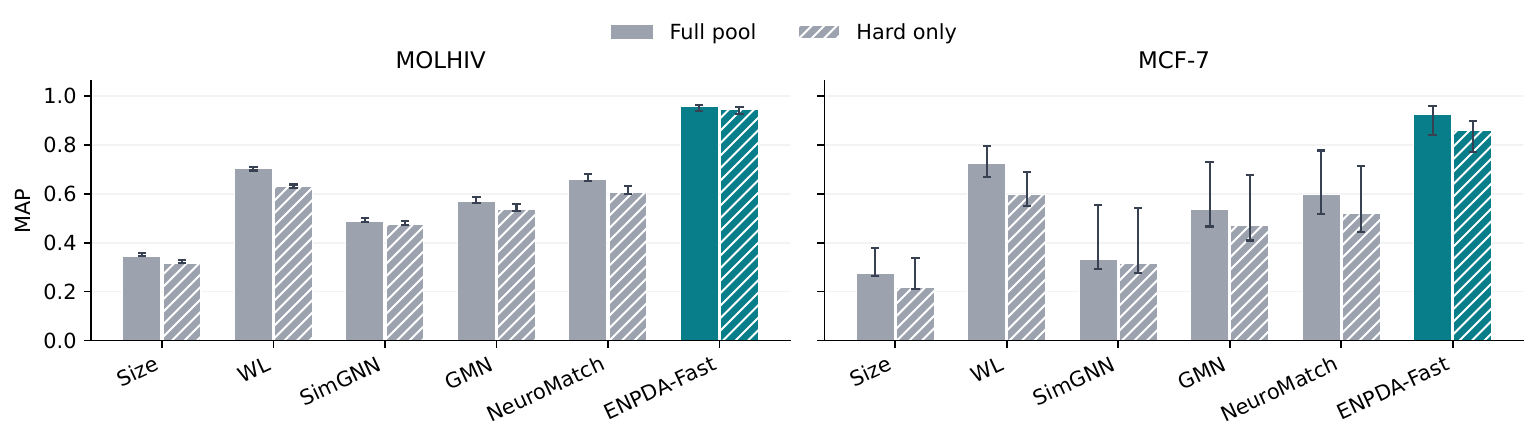}
\caption{Retrieval MAP on identical candidate pools.
Bars average three pools. Error bars bound unresolved oracle-label uncertainty.
Hard-only results exclude the controlled positive variants.}
\label{fig:enpda-retrieval}
\end{figure}

Each of three fixed pool seeds (20260823--20260825) defines 20 queries and 500 candidates per
query, for 30,000 pairs per dataset on MOLHIV and MCF-7.  Pool construction excludes the union of all relevant
reference, student and supervised-baseline training/validation graph identities, using exact
label-preserving isomorphism after WL prefiltering.  No model predictions or oracle labels
enter query, parent or candidate selection.  Within each pool, all six methods use the same
identities.  The neural baselines retain their independently trained supervised checkpoints.

Each query has eight controlled edited variants and 492 hard candidates from disjoint parents.
Hardness uses $0.65$ WL cosine plus $0.35$ size upper bound.  The fixed construction draws from
650 parents and up to two derived candidates per parent.  For MOLHIV pool seed 20260825,
one query exhausted the initial variant attempts after seven accepted candidates; deterministic
continuation with additional fixed seeds supplied the eighth, preserving the query and the first
seven variants.  It used the same edit types and counts (2--9), with acceptance at continuation
level 44.  This construction extension was completed before labels or predictions and is recorded
in the manifest; the six pools do not have identical attempted-generation budgets.

RASCAL with a 60-second per-pair cap supplies reference similarity intervals.
We follow the released NGA evaluator~\citep{ying2026neural}: MRR averages the reciprocal
predicted rank of the candidate with maximum reference similarity, choosing the first candidate
in the original order when reference similarities tie. P@10 and MAP define relevance as
reference similarity strictly greater than 0.5. Unresolved labels remain unknown rather than
becoming negatives.
Exact prefixes and unresolved tie blocks produce lower and upper MRR, P@10 and MAP bounds;
these are label-uncertainty bounds on the fixed pools, not sampling confidence intervals.
We average query metrics within each pool and then average the three pools.  Model seed 0/1/2
is paired with pool seed 20260823/24/25, so the aggregate combines model and pool variation.
There are 406 unresolved oracle pairs on MOLHIV and 5,339 on MCF-7, all retained.

ENPDA-Fast uses one learned four-round trajectory, four discrete restarts and four constructive
restarts, with at most ten local-refinement passes.  It uses no analytic trajectory stream,
annealing or large-neighborhood search.  Its returned mapping's Johnson similarity determines
rank.  The five baselines are size, WL, SimGNN, GMN and NeuroMatch.

\begin{table}[htbp]
\centering\small
\caption{Full-pool retrieval over three independently constructed pools. Entries are point estimates with bounds due to unresolved oracle labels, not confidence intervals. All six methods use identical query/candidate identities.}
\label{tab:retrieval-main}
\setlength{\tabcolsep}{4pt}\renewcommand{\arraystretch}{1.04}
\begin{tabular}{lrrrr}
\toprule
Dataset & Method & MRR [bounds] & P@10 [bounds] & MAP [bounds]\\
\midrule
MOLHIV & Size & 0.026 \confint{0.026}{0.026} & 0.392 \confint{0.392}{0.402} & 0.345 \confint{0.345}{0.356}\\
MOLHIV & WL & 0.375 \confint{0.375}{0.375} & 0.980 \confint{0.980}{0.980} & 0.704 \confint{0.695}{0.709}\\
MOLHIV & SimGNN & 0.020 \confint{0.020}{0.020} & 0.442 \confint{0.442}{0.455} & 0.485 \confint{0.484}{0.500}\\
MOLHIV & GMN & 0.152 \confint{0.152}{0.152} & 0.688 \confint{0.688}{0.705} & 0.569 \confint{0.563}{0.588}\\
MOLHIV & NeuroMatch & 0.233 \confint{0.233}{0.233} & 0.763 \confint{0.763}{0.778} & 0.657 \confint{0.651}{0.680}\\
MOLHIV & ENPDA-Fast & \textbf{0.975} \confint{0.975}{0.975} & \textbf{0.998} \confint{0.998}{0.998} & \textbf{0.957} \confint{0.940}{0.962}\\
MCF-7 & Size & 0.032 \confint{0.032}{0.034} & 0.420 \confint{0.420}{0.470} & 0.273 \confint{0.262}{0.379}\\
MCF-7 & WL & 0.437 \confint{0.437}{0.450} & 0.963 \confint{0.963}{0.965} & 0.723 \confint{0.668}{0.794}\\
MCF-7 & SimGNN & 0.013 \confint{0.013}{0.013} & 0.237 \confint{0.237}{0.388} & 0.330 \confint{0.291}{0.553}\\
MCF-7 & GMN & 0.139 \confint{0.136}{0.152} & 0.637 \confint{0.637}{0.723} & 0.534 \confint{0.466}{0.731}\\
MCF-7 & NeuroMatch & 0.335 \confint{0.334}{0.348} & 0.683 \confint{0.683}{0.722} & 0.597 \confint{0.516}{0.777}\\
MCF-7 & ENPDA-Fast & \textbf{0.972} \confint{0.971}{0.983} & \textbf{0.990} \confint{0.990}{0.990} & \textbf{0.926} \confint{0.840}{0.959}\\
\bottomrule
\end{tabular}
\end{table}

\begin{table}[htbp]
\centering\small
\caption{Hard-only retrieval over three independently constructed pools. Entries are point estimates with bounds due to unresolved oracle labels, not confidence intervals. All six methods use identical query/candidate identities.}
\label{tab:retrieval-hard}
\setlength{\tabcolsep}{4pt}\renewcommand{\arraystretch}{1.04}
\begin{tabular}{lrrrr}
\toprule
Dataset & Method & MRR [bounds] & P@10 [bounds] & MAP [bounds]\\
\midrule
MOLHIV & Size & 0.015 \confint{0.014}{0.026} & 0.305 \confint{0.305}{0.325} & 0.316 \confint{0.316}{0.329}\\
MOLHIV & WL & 0.414 \confint{0.405}{0.418} & 0.842 \confint{0.842}{0.847} & 0.630 \confint{0.623}{0.638}\\
MOLHIV & SimGNN & 0.015 \confint{0.015}{0.029} & 0.442 \confint{0.442}{0.455} & 0.474 \confint{0.473}{0.490}\\
MOLHIV & GMN & 0.146 \confint{0.145}{0.149} & 0.632 \confint{0.632}{0.648} & 0.537 \confint{0.531}{0.557}\\
MOLHIV & NeuroMatch & 0.147 \confint{0.143}{0.158} & 0.665 \confint{0.665}{0.682} & 0.606 \confint{0.600}{0.632}\\
MOLHIV & ENPDA-Fast & \textbf{0.776} \confint{0.709}{0.788} & \textbf{0.962} \confint{0.962}{0.970} & \textbf{0.946} \confint{0.928}{0.954}\\
MCF-7 & Size & 0.047 \confint{0.031}{0.153} & 0.233 \confint{0.233}{0.328} & 0.216 \confint{0.210}{0.336}\\
MCF-7 & WL & 0.496 \confint{0.432}{0.610} & 0.773 \confint{0.773}{0.812} & 0.597 \confint{0.550}{0.689}\\
MCF-7 & SimGNN & 0.034 \confint{0.026}{0.099} & 0.232 \confint{0.232}{0.383} & 0.313 \confint{0.275}{0.542}\\
MCF-7 & GMN & 0.156 \confint{0.121}{0.238} & 0.563 \confint{0.563}{0.660} & 0.472 \confint{0.409}{0.678}\\
MCF-7 & NeuroMatch & 0.187 \confint{0.139}{0.321} & 0.597 \confint{0.597}{0.665} & 0.520 \confint{0.443}{0.714}\\
MCF-7 & ENPDA-Fast & \textbf{0.694} \confint{0.600}{0.822} & \textbf{0.923} \confint{0.923}{0.927} & \textbf{0.858} \confint{0.770}{0.899}\\
\bottomrule
\end{tabular}
\end{table}

Full-pool ENPDA lower bounds exceed every baseline upper bound for all three metrics on both
datasets (Figure~\ref{fig:enpda-retrieval}).  Hard-only P@10 and MAP retain this ordering; MCF-7 hard-only MRR overlaps WL
(ENPDA lower 0.600 versus WL upper 0.610), so that comparison remains unresolved.

\begin{table}[htbp]
\centering\small
\caption{Complete-pool retrieval timing: evaluation wall seconds divided by 20 queries is batch throughput at eight workers, not serial-query latency. Worker seconds sum per-pair work. A dash marks an interrupted run whose final marker covers only the resumed suffix.}
\label{tab:retrieval-timing}
\setlength{\tabcolsep}{4pt}\renewcommand{\arraystretch}{1.04}
\begin{tabular}{lrrrr}
\toprule
Dataset & Seed & Timeouts / 10k & Wall s/query & Worker s/query\\
\midrule
MOLHIV & 0 & 106 & 57.78 & 456.55\\
MOLHIV & 1 & 222 & 64.96 & 513.90\\
MOLHIV & 2 & 78 & 65.22 & 515.30\\
MCF-7 & 0 & 1448 & --- & 974.44\\
MCF-7 & 1 & 1831 & 132.90 & 1057.14\\
MCF-7 & 2 & 2060 & 148.17 & 1179.59\\
\bottomrule
\end{tabular}
\end{table}

Timing in Table~\ref{tab:retrieval-timing} divides an eight-worker evaluation's full wall time by
20 queries.  It measures batch throughput, including refinement, rather than serial request
latency; summed worker seconds measure work, not wall time.  MCF-7 seed 0 was interrupted after
8,421 predictions, then resumed its 1,579 missing records.  Its final completion marker covers
only that suffix and is not a complete-pool wall-time measurement.  Combining the original
start-to-stop/delete window with the resumed evaluation gives 120.84--121.29 seconds/query,
and a three-seed average of 133.97--134.12 under that accounting.  This reconstruction excludes
downtime, includes interrupted work and termination overhead, and is neither an uninterrupted
measurement nor a confidence interval.

\subsection{Reproducibility and audit coverage}
\label{app:reproducibility}

Frozen manifests record configurations, graph identities, checkpoint hashes and raw per-pair
outputs.  The consolidated report verifies all completed training/evaluation markers and keeps
independent inference measurements separate.  Direct scoring checks injection, label-compatible
preserved edges, their endpoint vertices and Johnson similarity against raw input graphs.
If auxiliary assignments match incompatible labels but preserve no edge, restricting to
preserved-edge endpoints gives the compatible witness without changing the score.
No failed pair or timeout is dropped because of its outcome.

The main native/control/search audit directly rescores 22,116 maps; auxiliary controls and
protein transfer add 6,692, and deterministic reevaluation adds 4,365.  The retrieval and D\&D
checks cover 60,000 and 75 mappings, respectively.  These are records across arms and seeds,
not counts of independent graph pairs.  Round and IMDB-planted raw files retain scores and
provenance but not mappings; their hashes, checkpoint identities, coverage and score algebra
are checked, while direct map rescoring is limited to records that store a mapping.
Similarity errors are calculated from verified scores; signed intervals and all oracle bounds
are retained in the tables and machine-readable evidence.

The code checks equivariance, projection feasibility, ACG scoring and repaired-price weak
duality on enumerated small instances.  These enumerated checks do not replace the
exact-arithmetic guarantees.  Source code,
configuration manifests, raw outputs, audit records and rendering scripts form the reproducibility
artifact.  Graph-disjoint split and full cost accounts refer to the fresh models reported here.

Generative artificial intelligence (AI) tools assisted with language editing and code review.
The authors specified the method, inspected the mathematical statements, executed and verified
the experiments, and take responsibility for the manuscript and artifact.

\end{document}